\documentclass[mnsc,nonblindrev]{informs3_hide}
\OneAndAHalfSpacedXI 

\usepackage{graphicx} 
\usepackage[top = 1 in, bottom = 1.2 in, left = 1 in, right = 1 in]{geometry}
\usepackage{float}
\usepackage{booktabs}
\usepackage{siunitx}
\usepackage{amsfonts}
\usepackage{natbib}
\usepackage[table]{xcolor}
\usepackage[most]{tcolorbox}
\usepackage{booktabs}
\usepackage{array}
\usepackage[colorlinks,linkcolor=blue, citecolor=blue]{hyperref}
\usepackage{enumitem}
\usepackage{makecell}
\usepackage{subcaption}
\usepackage{graphicx}
\usepackage{float} 

\newcolumntype{P}{>{\raggedright\arraybackslash}p{0.22\textwidth}}
\newcolumntype{V}{>{\raggedright\arraybackslash}p{0.28\textwidth}}
\newcolumntype{U}{>{\raggedright\arraybackslash}p{0.18\textwidth}}
\newcolumntype{T}{>{\raggedright\arraybackslash}p{0.12\textwidth}}

\usepackage{amsmath,dsfont}

\usepackage{upgreek}
\usepackage{esvect}
\usepackage{cleveref}

\newcommand{\policy}{\ensuremath{\uppi}}
\newcommand{\paratheta}{\ensuremath{\theta}}
\newcommand{\policytheta}{\ensuremath{\policy_{\paratheta}}}

\newcommand{\prompt}{\ensuremath{x}}
\newcommand{\response}{\ensuremath{y}}

\newcommand{\Data}{\ensuremath{\mathcal{D}}}

\usepackage{natbib,bbm,multirow,multicol}
 \bibpunct[, ]{(}{)}{,}{a}{}{,}%
 \def\bibfont{\small}%
\TheoremsNumberedThrough     
\ECRepeatTheorems

\EquationsNumberedThrough    

\allowdisplaybreaks

\begin{document}



\RUNAUTHOR{}

\RUNTITLE{}

\TITLE{
Ask, Clarify, Optimize: Human–LLM Agent Collaboration for Smarter Inventory Control
}

\ARTICLEAUTHORS{%
\AUTHOR{Yaqi~Duan}

\AFF{Leonard N. Stern School of Business, New York University\  \\
\texttt{yaqi.duan@stern.nyu.edu}
}

\AUTHOR{Yichun~Hu}

\AFF{Johnson Graduate School of Management, Cornell University\  \\
\texttt{yh767@cornell.edu}
}

\AUTHOR{Jiashuo~Jiang}

\AFF{Hong Kong University of Science and Technology\  \\
\texttt{jsjiang@ust.hk}
}
}

\ABSTRACT{
 Inventory management remains a challenge for many small and medium-sized businesses that lack the expertise to deploy advanced optimization methods. This paper investigates whether Large Language Models (LLMs) can help bridge this gap. We show that employing LLMs as direct, end-to-end solvers incurs a significant ``hallucination tax'': a performance gap arising from the model's inability to perform grounded stochastic reasoning. To address this, we propose a hybrid agentic framework that strictly decouples semantic reasoning from mathematical calculation. In this architecture, the LLM functions as an intelligent interface, eliciting parameters from natural language and interpreting results while automatically calling rigorous algorithms to build the optimization engine.
 
To evaluate this interactive system against the ambiguity and inconsistency of real-world managerial dialogue, we introduce the Human Imitator, a fine-tuned ``digital twin'' of a boundedly rational manager that enables scalable, reproducible stress-testing. Our empirical analysis reveals that the hybrid agentic framework reduces total inventory costs by $32.1\%$ relative to an interactive baseline using \emph{GPT-4o} as an end-to-end solver. Moreover, we find that providing perfect ground-truth information alone is insufficient to improve \emph{GPT-4o}'s performance, confirming that the bottleneck is fundamentally computational rather than informational. Our results position LLMs not as replacements for operations research, but as natural-language interfaces that make rigorous, solver-based policies accessible to non-experts.

}

\KEYWORDS{Large Language Models, Inventory Management, Agentic Workflow, ERP Design, Decision Support Systems}


\maketitle

\section{Introduction}
\label{sec:intro}

A recurring tension in management science lies between \emph{normative optimality}—the decisions prescribed by formal models—and \emph{descriptive reality}—the decisions practitioners actually make under time pressure, incomplete data, and limited analytical support. Inventory management makes this tension vivid. Operations research has produced a mature toolkit for stochastic inventory control, including classical base-stock and $(s,S)$ policies with sharp optimality guarantees under stylized assumptions (e.g. \cite{axsater2006inventory}), as well as modern approximate dynamic programming and deep reinforcement learning (DRL) methods for richer environments (e.g. \cite{madeka2022deep}). In principle, these methods can reduce total cost (holding, ordering, and shortage) while improving service levels. In practice, however, many small and medium-sized retailers do not deploy them. Without dedicated analysts or data infrastructure, managers often rely on informal rules of thumb—``order when it looks low,'' ``double before holidays,'' ``round to a case-pack''—that are easy to apply but rarely calibrated to demand uncertainty, lead-time variability, or cost trade-offs.

This gap persists not primarily because the \emph{optimal policy} is too hard to compute, but because standard optimization tools remain inaccessible to non-experts. A critical challenge lies in the intricacy of \emph{problem formulation}. Even a basic inventory model requires translating messy operational context into precise inputs: what constitutes a stockout (lost sales vs.\ backorders); how lead times behave (deterministic vs.\ stochastic, including supply disruptions); what the review cadence is; which constraints bind (cash, storage, minimum order quantities, case-pack rounding, delivery schedules); and what objective proxies are acceptable (service-level targets, penalty costs, fill-rate). Much of this information is not stored in clean tables—it lives in human memory and informal language. This highlights an interpretive barrier: converting qualitative business narratives into a structured, model-consistent specification, and then converting the model output back into an operational routine that a manager can execute and trust.

Recent advances in large language models (LLMs) create a compelling opportunity to reduce this bottleneck, but also introduce a subtle trap. A natural first idea is to \emph{directly call the LLM} as an end-to-end decision-maker: describe the business in natural language and ask, “What should I order?” This often produces fluent, confident recommendations. Yet fluency is not the same as correctness. Inventory control is a stochastic control problem where small structural mistakes—misinterpreting lead time, conflating lost sales with backorders, mishandling uncertainty, ignoring capacity or service constraints—compound into recurring costs. Moreover, LLMs are not designed to guarantee feasibility, respect Bellman optimality, or provide calibrated probabilistic reasoning. As a result, a direct LLM call can yield policies that are \emph{plausible but systematically suboptimal}, or even internally inconsistent. In other words, naïvely using an LLM as the solver risks paying a “hallucination tax”: the efficiency loss induced when decision quality is limited by unconstrained language-model reasoning rather than grounded optimization.

The key insight of this paper is therefore not that LLMs can replace operations research, but that they can be used \emph{more intelligently}: as an interface and orchestrator that makes rigorous optimization usable. We propose a hybrid, agentic decision-support framework that explicitly separates semantic interaction from mathematical optimization. Our system decomposes the pipeline into three roles:
(i) a \emph{Information Extraction Agent} that engages the user, surfaces missing information, and resolves ambiguity through targeted follow-up questions;
(ii) an \emph{Optimization Agent} that receives a structured specification and computes a policy using grounded operations research algorithms (e.g., $(s,S)$-style methods and DRL where appropriate);
and (iii) an \emph{Policy Interpretation Agent} that translates the computed policy back into operational guidance—what to order and when—while explaining assumptions, checking feasibility against stated constraints, and presenting actionable summaries in the user’s own language.

This separation is not merely an engineering choice; it is an epistemological one. It treats language models as tools for elicitation, structuring, and translation, while reserving {policy computation} for methods that can be verified, stress-tested, and improved with established theory. This design also clarifies what the “last-mile” problem in decision support really is. When strong algorithms already exist, the binding constraint is often the {fidelity of the interface}: how well it can elicit, stabilize, and formalize the problem instance that the solver requires. In our framework, the LLM acts as a {rationality prosthetic}—a front end that sanitizes inputs, makes assumptions explicit, and aligns business narratives with model structure—while the optimization back end remains the engine of efficiency. Rather than pursuing a one-size-fits-all, end-to-end Generative AI solver, we argue for a dual-engine approach in which LLMs unlock the rigor of management science for users who cannot otherwise access it.

To rigorously evaluate this dual-engine architecture, however, we face a methodological challenge.
Interactive decision-support systems are difficult to evaluate at scale because real user inputs are noisy, inconsistent, and expensive to collect repeatedly. Static benchmarks miss the central difficulty: the agent must ask for missing parameters, handle contradictions, and converge to a well-posed model through dialogue. To enable controlled, reproducible evaluation, we introduce a \emph{Human Imitator}: a language model fine-tuned on more than hundreds of real human--machine dialogues. The Human Imitator serves as a scalable “digital twin” of a boundedly rational small-business owner, reproducing the ambiguity, incompleteness, and occasional inconsistency characteristic of real managerial inputs. This allows systematic stress-testing of interactive systems without the logistical and financial burden of large human-subject trials.

\subsection{Our Main Contributions}

Our study makes four main contributions to the design and evaluation of LLM-based decision support for stochastic inventory control.

    \begin{description}
        \item[Performance quantification:] Our hybrid agentic system reduces policy costs by $32.1\%$ relative to a baseline in which \emph{GPT-4o} is used as an interactive end-to-end solver. This gap provides a concrete estimate of the \emph{hallucination tax}: the efficiency loss incurred when policy computation relies on unconstrained language-model reasoning rather than grounded stochastic optimization.

        \item[Drivers of Performance:] We disaggregate performance to identify the specific regimes where agentic support yields the highest marginal value. Our analysis reveals that while the framework is distribution-agnostic (robust to demand shape), its advantage scales with \textit{complexity} and \textit{economic stakes}. The performance gap widens significantly under longer lead times and in high-penalty, high-flexibility regimes. This observed ``complexity premium'' confirms that solver-backed architectures deliver outsized returns precisely where the financial consequences of imprecise heuristics are most acute.

    \item[Limits of prompt-based reasoning:] We isolate the source of decision error by comparing the interactive end-to-end \emph{GPT-4o} baseline against a ``perfect information'' counterfactual. Strikingly, providing ground-truth parameters to \emph{GPT-4o} yields no statistical improvement. This identifies a hard ``cognitive ceiling'': the performance bottleneck of LLMs is not \textit{informational} (data extraction) but fundamentally \textit{computational}. This confirms that prompt engineering cannot bridge the gap to stochastic optimality; rather, LLMs function better when architected to \emph{orchestrate} rigorous solvers.

        \item[Behavioral simulation for interactive benchmarks:] Methodologically, we address the scarcity of interactive benchmarks by establishing a Human Imitator as a scalable proxy for boundedly rational managers. By reproducing the inconsistency and ambiguity of real-world inputs, this approach allows us to move beyond static datasets and stress-test decision support systems against the realistic friction of human-machine interaction. Beyond inventory control, our scalable evaluation pipeline offers a general template for evaluating interactive LLM-based decision-support tools in other operational domains.

    \end{description}
    
    Collectively, these findings suggest that the true potential of Generative AI in operations lies in its ability to democratize expertise. By serving as an intelligent orchestration layer on top of existing analytical tools, LLMs can finally unlock the power of management science for the long tail of small business owners who have historically been left behind.


    \subsection{Related Work}
Our work sits at the intersection of three distinct streams of literature: stochastic inventory control (specifically Deep Reinforcement Learning approaches), the application of Large Language Models (LLMs) to optimization, and the evaluation of interactive dialogue systems through user simulation.

\subsubsection{Stochastic Inventory Control and Deep Reinforcement Learning.}
The theoretical foundations of inventory management are mature, anchored by the optimality of $(s, S)$ policies for single-echelon systems with linear costs \citep{scarf1960optimality} and their efficient computation \citep{axsater2006inventory}. Interested readers may refer to textbooks such as \cite{simchi2005logic} for further details. However, real-world complexities—such as lead-time variability, lost sales, and multi-echelon networks—often render exact dynamic programming intractable due to the curse of dimensionality.
To address these high-dimensional state spaces, recent scholarship has pivoted toward Deep Reinforcement Learning (DRL). DRL, when combined with deep neural networks, effectively addresses the high-dimensional state and action spaces inherent in inventory control, alleviating the curse of dimensionality. Early studies demonstrated the feasibility of RL in inventory control. For instance, \cite{oroojlooyjadid2022deep} utilized the Deep Q-network (DQN) to solve the beer distribution game, a widely studied supply chain simulation. Similarly, \cite{gijsbrechts2022can} employed the A3C algorithm to achieve heuristic-level performance, while \cite{stranieri2023comparing} benchmarked multiple DRL methods, including A3C, PPO, and vanilla policy gradient (VPG), for inventory problems. Recent advancements have extended DRL to a variety of inventory control scenarios, such as managing non-stationary uncertain demand \citep{dehaybe2024deep, park2023adaptive}, optimizing multi-product systems \citep{sultana2020reinforcement, selukar2022inventory}, and handling diverse product types \citep{meisheri2020using, meisheri2022scalable}. DRL has also been applied to complex supply chain structures, including multi-echelon systems \citep{wu2023distributional, alvo2023neural, liu2024reinforcement, stranieri2024combining}, one-warehouse multi-retailer networks \citep{kaynov2024deep}, and the stochastic capacitated lot-sizing problem \citep{van2023using}.

Despite these algorithmic advances, a deployment gap remains. These methods require formal mathematical modeling and hyperparameter tuning that are inaccessible to the average small-business manager. Our work does not seek to invent a new DRL algorithm; rather, we leverage these existing powerful solvers (the ``Optimization Agent'') and focus on the \emph{interface} required to make them usable by non-experts.

\subsubsection{LLMs for Optimization and Decision Support.}
The emergence of LLMs has sparked intense interest in automating decision making (e.g. \cite{huang2025orlm}. LLMs are increasingly being adopted as versatile decision-support tools across many critical sectors. Within business and operations management, LLMs are leveraged to optimize processes, improve efficiency, and drive innovation \citep{li2024large}. Significant recent research interest has focused on their ability to automatically formulate optimization problems and dynamic programming problems from natural language descriptions \citep{ahmaditeshnizi2023optimus,ahmaditeshnizi2024optimus,zhou2025auto,liang2025llm}. In supply chain management specifically, LLMs are used for tasks such as demand forecasting and logistics optimization \citep{lu2024optimizing}, and recent work has also focused on developing agentic frameworks for advanced applications \citep{quan2024invagent,qi2025leveraging,long2025autonomous,simchi2026large,cohen2025supply}. In the healthcare domain, LLMs show remarkable potential to augment clinical decision-making by rapidly synthesizing patient information, generating differential diagnoses, and suggesting treatment plans, positioning them as powerful co-pilots for clinicians \citep{shah2023creation,nazi2024large}. Similarly, the financial sector uses LLMs for a wide range of applications, from market analysis to risk management, with specialized models like FinGPT and BloombergGPT enabling nuanced sentiment analysis, automated report generation, and enhanced algorithmic trading strategies.

Our work advances this direction by addressing the \emph{ambiguity} of the inputs. Existing frameworks typically assume the user provides a complete problem description. In contrast, our \emph{Information Extraction Agent} assumes the user is boundedly rational and the problem is initially ill-posed, requiring an iterative dialogue to elicit necessary parameters (e.g., distinguishing backorders from lost sales) before the solver can be invoked. This design makes our framework particularly suitable for small and medium-sized enterprises that lack in-house analytical staff but still face nontrivial inventory trade-offs.

\subsubsection{Generative Agents and User Simulation.}
Evaluating interactive decision support systems presents a methodological dilemma: static datasets (e.g., standard NLP benchmarks) fail to capture the multi-turn dynamics of problem formulation, while human-subject trials are resource-intensive and difficult to reproduce. To address this, we draw upon the rich history of user simulation in dialogue systems and the recent emergence of generative agents.

User simulators have long been a staple in training task-oriented dialogue systems, particularly for Reinforcement Learning (RL) agents. Early approaches relied on agenda-based mechanisms \citep{schatzmann2007agenda}, where the simulator followed a strict stack of goals (e.g., ``book a flight'', ``specify time''). While effective for slot-filling tasks, these rule-based systems lacked the linguistic diversity of real users. The field subsequently moved toward data-driven approaches, utilizing sequence-to-sequence models to learn user behavior directly from corpora \citep{kreyssig2018neural, shi2019build}. However, these models often struggled with response collapse, generating generic or repetitive answers that failed to challenge the system.
The advent of Large Language Models has revolutionized user simulation by enabling ``Generative Agents'' that maintain consistent personas and memories. \cite{park2023generative} shows that LLMs could simulate believable social interactions in a sandbox environment. In the context of task-oriented dialogue, \cite{wang2025user} established that LLMs can act as zero-shot user simulators that outperform traditional models.
Crucially for our work, recent literature emphasizes simulating human imperfections. \cite{wang2024incharacter} investigated role-playing capabilities, finding that LLMs can effectively mimic specific demographic traits and knowledge gaps. \cite{li2023camel} introduced ``CAMEL'', a framework where two LLM agents (a user and an assistant) interact autonomously to solve tasks, revealing that agent-to-agent simulation can uncover edge cases that static testing misses.

Our \emph{Human Imitator} synthesizes these streams. Unlike the perfect ``agenda-based'' simulators of the past, our agent is designed to replicate the ``bounded rationality'' \citep{simon1955behavioral} of a non-expert supply chain manager. By conditioning the simulator on personas that exhibit specific knowledge gaps (e.g., confusing ``lost-sale'' with ``back-order''), we create a rigorous testbed that evaluates not just the system's ability to solve math, but its ability to clarify ambiguity. By fine-tuning a model on real human-machine interactions, we create a ``digital twin'' of a small business owner. This allows us to stress-test the system's ability to handle noise and ambiguity at scale, providing a more rigorous evaluation metric than simple code-generation accuracy.


    \section{The Hybrid Agentic Framework}
\label{sec:framework}

In this section, we present the design of our LLM-based solver. The core philosophy of our framework is the strict separation of concerns: rather than employing a single ``black box'' model, we architect a modular system where specialized agents handle distinct cognitive tasks: ambiguity resolution, mathematical optimization, and semantic interpretation.

This design enables us to bridge the gap between the ambiguous descriptions often found in real-world business and the precise inputs required by formal inventory models (formulated in \Cref{sec:InventoryModel}). By organizing these agents into a collaborative workflow, our system functions as a ``cognitive scaffold'': it standardizes human inputs for rigorous solving and subsequently interprets the mathematical results into actionable insights.

We begin by describing the structure of the agentic pipeline, followed by a detailed breakdown of each individual agent.

\vspace{-.5em}

\subsection{Pipeline Architecture}

\begin{figure}[!t]
		\centering
        \vspace{-.5em}
		\includegraphics[width = \linewidth]{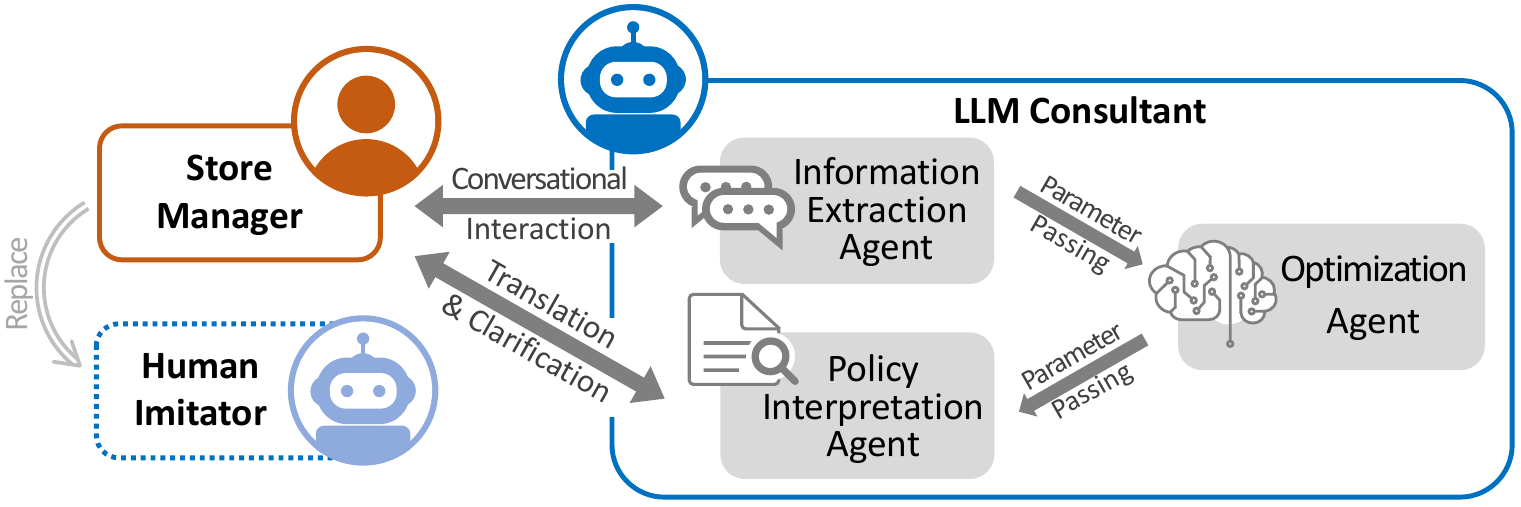} 
		\caption{LLM-Agent Pipeline for Inventory Management}
		\label{fig:pipeline}
	\end{figure}

Our framework integrates human-like problem input, structured information extraction, adaptive optimization, and verbal policy interpretation into a coherent pipeline (\Cref{fig:pipeline}).
	The pipeline is composed of three interconnected components that together transform informal descriptions into optimized inventory policies. 
    \begin{description}[leftmargin=4em, labelsep=1em] 
		\item[Information Extraction Agent.] This agent converses with the human user, interprets the user's informal input, and incrementally converts it into structured parameters. Through conversational rounds, it asks clarifying questions, fills missing entries, and resolves conflicts until a complete and consistent parameter table is obtained.
		\item[Optimization Agent.] Once the problem is fully specified, the Optimization Agent invokes appropriate solvers, such as those for the classical $(s, S)$ policy and deep reinforcement learning algorithms, to generate inventory policies. The choice of the recommended policy depends on user-defined preferences, e.g., trading off expected cost against cost variance.
        \item[Policy Interpretation Agent.] Once the Optimization Agent identifies a recommended policy, this policy is passed to the Policy Interpretation Agent, which intuitively explains the nature of the policy to the user, responds to optimal order quantity queries, and addresses any clarification questions.
		\end{description}
	Together, these components establish an iterative, human-like workflow that mimics real managerial practice while remaining fully automatable.

    \subsection{Information Extraction: Turning Dialogue into Parameters}

    When a user approaches our system for help with an inventory problem, the primary barrier to applying operations research methods is not optimization itself, but \emph{problem specification}: converting an informal, natural-language business description into a fully specified mathematical model. In our setting, the necessary information is provided incrementally through a multi-turn conversation, so the system must continuously extract, reconcile, and refine model parameters as the dialogue evolves. We assign this responsibility to the \textbf{Information Extraction Agent} (built on \emph{GPT-5-mini}\footnote{Unless otherwise noted, \emph{GPT-4o} serves as the baseline LLM throughout this paper. \emph{GPT-5-mini} is a smaller and weaker model; we use it here to reduce cost, as its performance is sufficient for the information-extraction task.}).

    In such conversational AI systems, relying solely on the model’s implicit memory---that is, the context window---is brittle for operations research applications. Long, unstructured dialogues are prone to \emph{context drift}, in which previously specified constraints are forgotten, and \emph{hallucination}, in which missing parameters are spuriously fabricated. To mitigate these failure modes, we adopt an \emph{Explicit Memory} architecture. Concretely, the agent is required to maintain a persistent \emph{Parameter Specification Table} (e.g., \Cref{tab:initial_params}) that represents the system’s authoritative epistemic state.
    
    The agent’s objective is then to populate this table by systematically eliciting, inferring, and validating all decision-relevant primitives. Through this process, the agent transforms an initially empty template (e.g., \Cref{tab:initial_params}) into a fully instantiated problem representation (e.g., \Cref{tab:completed_params}) that can be passed directly to the solver.

    \begin{table}[!ht]
        \centering
        \renewcommand{\arraystretch}{1.1}
        \begin{tabular}{P V U T}
            \toprule
            \textbf{Parameter} & \textbf{Value} & \textbf{Unit} & \textbf{Status} \\
            \midrule
            time\_horizon            & -- & --  & undefined \\
            demand\_type             & -- & N/A & undefined \\
            demand\_distribution     & -- & N/A & undefined \\
            perishability            & -- & N/A & undefined \\
            state\_transition\_model & -- & N/A & undefined \\
            holding\_cost            & -- & --  & undefined \\
            penalty\_cost            & -- & --  & undefined \\
            setup\_cost              & -- & --  & undefined \\
            lead\_time               & -- & --  & undefined \\
            max\_inventory           & -- & N/A & undefined \\
            max\_order               & -- & N/A & undefined \\
            risk\_tolerance          & -- & N/A & undefined \\
            \bottomrule
        \end{tabular}
        \caption{\textbf{Initial Parameter Specification for the Information Extraction Agent}}
        \vspace{1.2em}
        \label{tab:initial_params}
        
        \renewcommand{\arraystretch}{1.1}
        \begin{tabular}{P V U T}
            \toprule
            \textbf{Parameter} & \textbf{Value} & \textbf{Unit} & \textbf{Status} \\
            \midrule
            time\_horizon            & 60 & days & defined \\
            demand\_type             & random & N/A & defined \\
            demand\_distribution     & Poisson($\lambda=10$) & N/A & defined \\
            perishability            & TRUE & N/A & defined \\
            state\_transition\_model & lost\_sale & N/A & defined \\
            holding\_cost            & 5 & USD/unit/day & defined \\
            penalty\_cost            & 15 & USD/unit & defined \\
            setup\_cost              & 100 & USD/order & defined \\
            lead\_time               & 5 & days & defined \\
            max\_inventory           & 100 & N/A & defined \\
            max\_order               & 20 & N/A & defined \\
            risk\_tolerance          & 3 & N/A & defined \\
            \bottomrule
        \end{tabular}
        \caption{\textbf{Example of a Completed Parameter Specification}}
        \label{tab:completed_params}
    \end{table}

    \begin{figure}[!t]
		\centering
        \vspace{-.5em}
		\includegraphics[width = 0.8\linewidth]{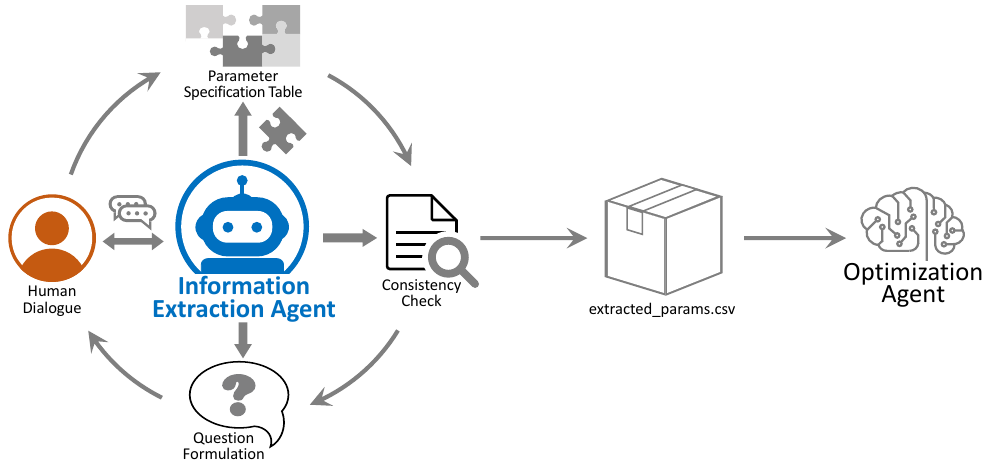} 
		\caption{Architecture of the Information Extraction Agent}
		\label{fig:info_extra}
	\end{figure}
    
    More concretely, the conversational process unfolds in rounds. In each round, the Information Extraction Agent analyzes the user's latest response and performs the following actions:
    \begin{enumerate}
        \item \emph{Update Table:} If new, credible information is identified, it fills the corresponding blank in the table and updates the status to ``defined.''
        \item \emph{Detect Conflicts:} If the user provides information that contradicts a previously defined entry, the agent flags the conflict for resolution.
        \item \emph{Formulate Question:} If the table contains undefined variables or identified conflicts, the agent generates a targeted question to clarify ambiguities or gather missing data. The conversation continues to the next round.
    \end{enumerate}
    The dialogue concludes when all parameters are marked as ``defined'' and the system verifies that no conflicts exist. A visual representation of this complete workflow is provided in \Cref{fig:info_extra}.
    For reproducibility, the exact system prompt used by the Information Extraction Agent is documented in \Cref{tab:system_prompt_info_extraction} (\Cref{sec:system_prompts}). Furthermore, we provide representative conversation logs in \Cref{sec:example1,sec:sample_conversation}; specifically, Examples 3--5 highlight the agent's capability to navigate ambiguity and resolve logical conflicts effectively.

    Upon successful termination, the Information Extraction Agent saves two files: 
    \begin{itemize} 
        \item \texttt{extracted\_params.csv}, containing the extracted parameters from the conversation; and 
        \item \texttt{conversation\_log.txt}, containing the complete conversation history. 
    \end{itemize}
    Finally, it invokes the Optimization Agent, which uses the \texttt{extracted\_params.csv} file to calculate and propose a recommended inventory policy.

\begin{remark}[Comparison with DPLM]
    With the above extraction-and-memory mechanism in place, we now situate our approach relative to recent efforts on automating operations-research modeling.
    We position our work by contrasting it with the recent Dynamic Programming Language Model (DPLM) proposed by \citet{zhou2025auto}. While sharing the ultimate goal of automating operations research modeling, our approaches differ fundamentally in their handling of information ambiguity and training methodology. \citet{zhou2025auto} demonstrate that a specialized fine-tuning pipeline trained on synthetic data can effectively translate fully specified, static textbook descriptions into mathematical models. In contrast, our framework addresses the \emph{preceding} challenge of real-world ambiguity, where problem definitions must be interactively elicited and maintained through unstructured dialogue. We demonstrate that for this extraction phase, rigorous prompt engineering with a capability-rich foundation model (\emph{GPT-5-mini}) and explicit memory management yields high-fidelity results without requiring the extensive supervised fine-tuning and reinforcement learning resources detailed in \citet{zhou2025auto}. However, their rigorous synthetic data generation (DualReflect) and fine-tuning recipes offer a valuable roadmap for future enhancements, particularly for distilling our extraction capabilities into smaller, more efficient models or for improving the system's ability to auto-formulate novel constraints that fall outside standard inventory templates.
\end{remark}

    \subsection{Optimization: From Parameters to Policies}

\begin{figure}[!t]
    \centering
    \includegraphics[width = \linewidth]{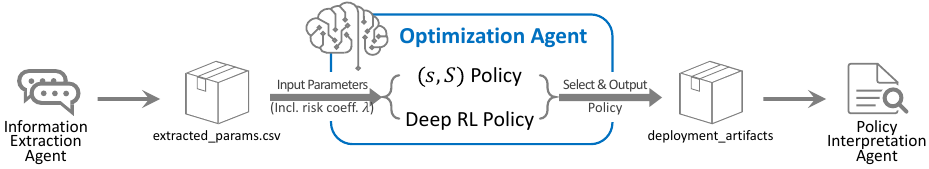} 
    \vspace{-2em}
    \caption{Architecture of the Optimization Agent}
    \label{fig:LLM_Solver}
\end{figure}

    With the problem structure specified, the \textbf{Optimization Agent} then executes the core algorithmic work. Unlike a generic LLM that might ``hallucinate'' a solution, this agent acts as a dispatcher for rigorous operations research solvers. The corresponding workflow is illustrated in Figure \ref{fig:LLM_Solver}.

    A key feature of our optimization module is the integration of user risk preferences. We recognize that normative optimality is not single-dimensional; users often trade off expected cost against stability. To capture this, the agent elicits a preference parameter $\lambda \in [-10, 10]$, which weights the penalty for variance (risk). The agent then selects the policy $\policy$ that minimizes the following composite objective:
    \begin{equation}\label{eqn:objective}
        \mathcal{L}(\policy) \; = \; \mathds{E}\big[\mathrm{Cost}(\policy)\big] \; + \; \exp(-\lambda) \, \cdot \, \mathrm{Std}\big[\mathrm{Cost}(\policy) \big] \, ,
    \end{equation}
    where $\mathrm{Cost}(\policy)$ represents the expected cost and $\mathrm{Std}(\policy)$ represents the cost standard deviation under policy $\policy$.

In our current implementation, the agent arbitrates between two distinct policy classes, representing the spectrum from classical theory to modern AI:

\begin{description}
    \item[The $(s, S)$ Policy (Classical Heuristic):] The $(s, S)$ policy is a classic inventory management approach defined by a reorder point $s$ and an order-up-to level $S$. When the inventory level (the sum of on-hand inventory and orders in the pipeline) drops to or below $s$, a replenishment order is triggered to bring the stock back up to $S$. This policy effectively balances ordering and holding costs by maintaining inventory within these thresholds, ensuring stock is replenished only when necessary. It is particularly effective in systems with positive lead times. While the policy is simple to implement and interpret, improper values for $s$ and $S$ can lead to significant overstocking or stockouts. Our Optimization Agent simulates the expected cost and variance across a grid of $(s, S)$ parameter choices to identify the best $(s, S)$ configuration.
    \item[Deep Q-Network (Deep Reinforcement Learning):] The DQN policy is based on the Markov Decision Process~(MDP) formulation of the inventory problem (detailed in \Cref{sec:InventoryModel}). It is a reinforcement learning method that combines Q-learning with deep neural networks to learn optimal policies in dynamic environments. Rather than maintaining a Q-table, which is infeasible for large state-action spaces, DQN uses a neural network to approximate the Q-value function, predicting the expected cumulative reward for each action given a state. We define the state space as the composition of the current inventory position, the time period, and the vector of orders in the pipeline. By iteratively updating the Q-network via the Bellman equation, DQN approximates optimal policies using samples from the demand distribution. Notably, DQN performs well even when the lead time is large due to its ability to handle sequential decision making problems with high-dimensional input.    
    \end{description}    
    These two paradigms represent an inherent trade-off between \textit{expected efficiency} and \textit{operational stability}. While the Deep RL policy leverages high-dimensional feature extraction to aggressively minimize expected costs, its stochastic nature may introduce higher variance. Conversely, the $(s, S)$ heuristic, though potentially conservative, offers a robust and predictable baseline.   
    Our framework addresses this dichotomy through an \textit{agentic design}: the Optimization Agent interprets the user's risk tolerance and performs meta-reasoning to arbitrate between the solvers, selecting a policy that balances expected cost and operational stability.

        Upon convergence, the Optimization Agent serializes the model artifacts to a directory named \texttt{deployment\_artifacts}, containing three files:
    
    \begin{itemize}
        \item \texttt{config.json}: Includes important instance parameters (e.g., lead time, maximum inventory, time horizon, and the state and action dimensions of the DQN network).
        \item \texttt{dqn\_weights.pth}: Contains the weights of the DQN network.
        \item \texttt{policy\_evaluation\_results.csv}: Stores the average cost and standard deviation of the optimal DQN policy and $(s, S)$ policy, respectively, along with the optimal $s$ and $S$ values.
    \end{itemize}
    These artifacts serve as the essential input for the subsequent Policy Interpretation Agent.

\begin{remark}[Extensibility to Novel Algorithms]
    Although our experiments focus on $(s,S)$ and DQN policies, the modular architecture of the Optimization Agent facilitates the seamless integration of additional policy classes and emerging algorithms. Unlike black-box AI models, where enhancing algorithmic reasoning often necessitates resource-intensive retraining or fine-tuning, our framework treats solvers as interchangeable modules. This design allows researchers to effortlessly incorporate novel optimization techniques, thereby ensuring the system remains at the frontier of operations research without altering the core linguistic interface.
\end{remark}

    \subsection{Policy Interpretation: Communicating Policies in Words}
    The final component, the \textbf{Policy Interpretation Agent}, ingests the artifacts produced by the Optimization Agent and translates them into actionable business intelligence. This stage is critical for realizing actual cost reductions: theoretical optimality translates to practical efficiency only when human decision-makers trust and adhere to the algorithmic suggestions.\footnote{Pre-LLM studies of automated replenishment already document that store managers frequently deviate from optimization-based order recommendations and that accounting for their behavioral responses is essential for effective system design \citep{van2010ordering}.} 
    To mitigate algorithm aversion and foster cognitive alignment, this agent navigates the trade-off between \emph{performance} (lower cost $\mathcal{L}(\policy)$ in \cref{eqn:objective}) and \emph{interpretability}.

    Because the $(s, S)$ policy is generally more interpretable and familiar to practitioners, while the DQN policy is less transparent but typically achieves lower costs, the Decision Interpretation Agent follows the communication protocol below:

    \begin{description}
        \item[\textbf{Scenario A:} The $(s, S)$ Policy Performs Better.] 
If the $(s, S)$ heuristic yields a lower objective value $\mathcal{L}(\policy)$, the agent leverages its inherent interpretability. It provides a detailed yet accessible explanation of the reorder point $s$ and order-up-to level~$S$, describing how the policy operates in day-to-day inventory decisions. It then invites the user to ask follow-up questions or relate the thresholds to their own operational experience.

        \item[\textbf{Scenario B:} The DQN Policy Performs Better.] If the DQN policy yields a lower objective value~$\mathcal{L}(\policy)$, the agent first offers a high-level overview of the black-box policy, framing it as an ``on-demand expert’’: at any time step, given the current inventory position and pipeline orders, the user can query the agent for a recommended order quantity. If the user finds this lack of interpretability unsatisfying and prefers a more intuitive explanation, the agent then additionally presents the suboptimal $(s, S)$ policy, reports the percentage difference in the optimization objective between the two policies (e.g., ``If you're interested in a simpler policy, the $(s=2, S=11)$ policy exists but it costs 20.5\% more''), and allows the user to decide which policy they ultimately wish to adopt.
        
    \end{description}

\begin{figure}[!t]
    \centering
    \includegraphics[width = \linewidth]{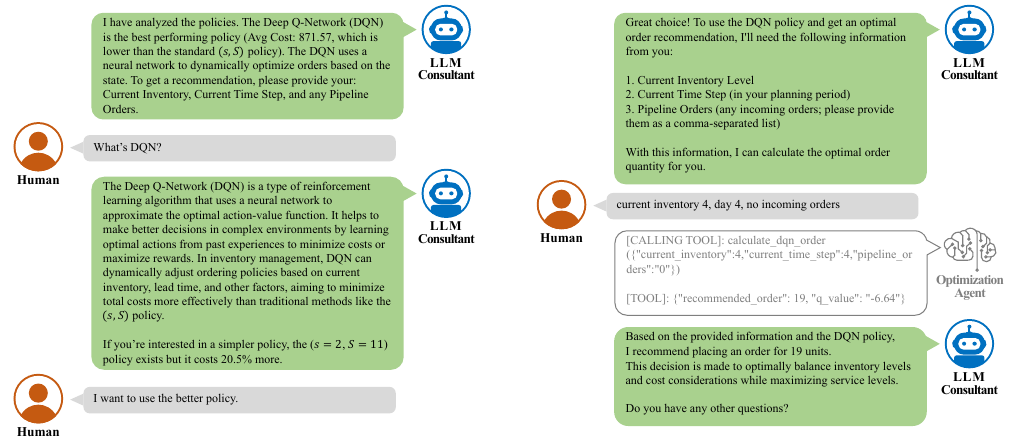} 
    \vspace{-2em}
    \caption{Representative dialogue between the Human User and the Policy Interpretation Agent.}
    \label{fig:interpretation}
\end{figure}

    Throughout this interaction, the agent remains responsive to user inquiries, clarifying the rationale behind specific recommendations to reinforce trust. Once the user indicates understanding or signals readiness to adopt the policy, the agent terminates the conversation and archives the dialogue transcript as \texttt{interpret.txt} for audit and analysis. An illustrative example of this interaction is provided in \Cref{fig:interpretation}.

\section{The Human Imitator: A Scalable Testbed}
\label{sec:imitator}

To rigorously evaluate our agentic framework presented in \Cref{sec:framework}, we must confront a fundamental methodological challenge: the scarcity of realistic testbeds. Standard inventory benchmarks rely on static, structured parameters, failing to capture the ambiguity, inconsistency, and bounded rationality that define real-world decision-making. Conversely, human-subject experiments, while ecologically valid, are costly, unscalable, and difficult to reproduce.

We bridge this gap by developing the \emph{Human Imitator}—an LLM fine-tuned to function as a ``digital twin'' of a retail manager. Unlike a standard LLM, which strives for helpfulness and precision, our imitator is explicitly trained to reproduce the idiosyncrasies of human behavior: providing rough estimates, using informal language, and occasionally omitting critical details. This allows us to stress-test our system against realistic ``human noise'' at scale.

\subsection{Data Collection: Capturing Managerial Dialogue}

We constructed a specialized corpus of inventory management dialogues through a controlled study. We deployed an online interface where participants played the role of a store manager interacting with a generic ``Base Model'' consultant. The Base Model was prompted to elicit specific operational parameters—including demand patterns, perishability, holding costs, and lead times—while the participants were instructed to describe their business scenario using natural language.

This process yielded a dataset of prompt–response pairs, $\mathcal{D} = \{ (x^{(i)}, y^{(i)} \}_{i=1}^N$, where $x^{(i)}$ represents the system's inquiry (e.g., \emph{``How long do items stay fresh?''}) and $y^{(i)}$ represents the human's verbatim response (e.g., \emph{``maybe 3 days max''}). 

In total, $66$ participants generated $N = 1,184$ high-quality conversational turns. This dataset captures a diverse range of linguistic styles, from precise specifications to vague, heuristic-driven descriptions. Detailed screenshots of the interface and example transcripts of the collected dialogues are provided in \Cref{sec:SFT}.

\subsection{Supervised Fine-Tuning (SFT)}

To instill human-like behavioral patterns into the model, we performed Supervised Fine-Tuning (SFT) on the \emph{Qwen2.5-7B} foundation model. We treat the imitator as a policy $\policytheta$ parameterized by weights $\paratheta$. The training objective is to maximize the likelihood of generating the empirically observed human response $\response$ given the context $\prompt$:
\[
\max_\theta \quad \sum_{(\prompt^{(i)}, \response^{(i)}) \in \Data} \log \policytheta \big(\response^{(i)} \bigm| \prompt^{(i)} \big).
\]
This objective forces the model to align its probability distribution with human behavior, effectively teaching it to ``unlearn'' robotic precision and adopt the persona of a boundedly rational manager.

Since our goal is \emph{imitation}—creating a faithful proxy that reproduces the specific distribution of our target population—we train on the full dataset to maximize behavioral coverage. Training was conducted using Low-Rank Adaptation (LoRA) on a single NVIDIA L4 GPU (24GB). The optimization converged rapidly over 3 epochs (524 minutes), with the loss function reducing from $1.9234$ to $0.3473$, indicating that the foundation model successfully adapted to the domain-specific linguistic patterns. See \Cref{sec:SFTtraining} for experimental details. 

\subsection{Empirical Validation: Quantifying Behavioral Alignment}

\begin{figure}[!ht]
    \centering
    \includegraphics[width = 0.75 \linewidth]{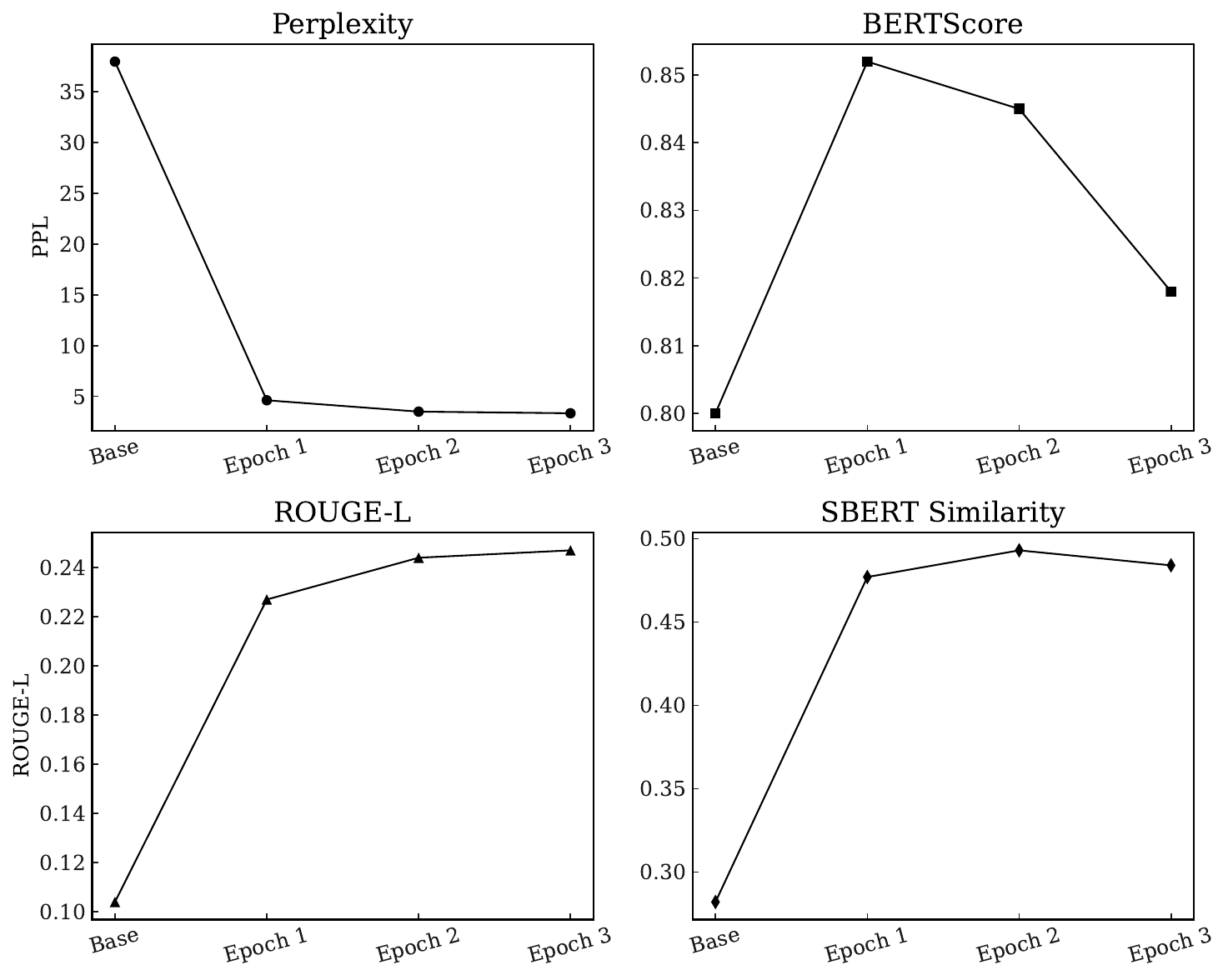} 
    \caption{\textbf{SFT Evaluation Trajectory.} The curves illustrate the rapid convergence of the Human Imitator. \\
    The drop in Perplexity and the simultaneous rise in semantic similarity metrics (BERTScore, ROUGE-L and SBERT) indicate successful domain adaptation.}
    \label{fig:SFT_eval}
\end{figure}

Does the fine-tuned model truly resemble a human manager? We monitor the model's fidelity across four metrics that measure alignment at distinct linguistic levels: \emph{fluency}, \emph{structure}, \emph{token usage}, and \emph{sentence meaning}. The training trajectory is visualized in \Cref{fig:SFT_eval}, with numerical results detailed in \Cref{tab:sft-eval}.

\begin{table}[ht]
    \centering
    \renewcommand{\arraystretch}{1.2}
    \begin{tabular}{l S[table-format=2.3] S[table-format=1.3] S[table-format=1.3] S[table-format=1.3]}
        \toprule
        \textbf{Training Stage} & \textbf{PPL} $\downarrow$ & \textbf{BERTScore} $\uparrow$ & \textbf{ROUGE-L} $\uparrow$ & \textbf{SBERT} $\uparrow$ \\
        \midrule
        Epoch 0 (Base) & 37.991 & 0.800 & 0.104 & 0.282 \\
        Epoch 1        &  4.615 & \textbf{0.852} & 0.227 & 0.477 \\
        Epoch 2        &  3.487 & 0.845 & 0.244 & \textbf{0.493} \\
        Epoch 3        &  \textbf{3.334} & 0.818 & \textbf{0.247} & 0.484 \\
        \bottomrule
    \end{tabular}
    \caption{\textbf{Validation of the Human Imitator.} We track four metrics across training epochs. \emph{PPL} measures predictive certainty (lower is better). \emph{BERTScore} measures token-level embedding similarity. \emph{ROUGE-L} measures structural Longest Common Subsequence (LCS) overlap. \emph{SBERT} measures sentence-level semantic alignment.}
    \label{tab:sft-eval}
\end{table}

\paragraph{Drastic Reduction in Perplexity (Domain Adaptation):}
The most critical indicator of success is Perplexity (PPL), which measures the model's predictive uncertainty regarding the next token. As shown in \Cref{tab:sft-eval}, the PPL drops precipitously from $37.99$ (Base~Model) to $3.33$ (Epoch~3). This order-of-magnitude improvement confirms that the model, which initially spoke like a generic assistant, has successfully adopted the specific jargon, hesitation, and vernacular of human input.

\paragraph{Semantic and Structural Convergence:}
To ensure the model captures the \emph{intent} and \emph{style} of the managers, we examine similarity metrics at three levels of abstraction:
\begin{itemize}
    \item \emph{BERTScore (Token-Level Embedding Similarity):} BERTScore computes the cosine similarity between contextual embeddings of individual tokens. A high score (e.g. $0.85$) confirms that the model is selecting the correct \emph{vocabulary} within context, ensuring granular alignment with the specific terminology used by inventory managers.
    \item \emph{ROUGE-L (Structural Similarity):} ROUGE-L measures the Longest Common Subsequence (LCS) between the generated and reference text. The significant rise (from 0.104 to 0.247) indicates that the imitator is replicating the \emph{syntactic structure} of the participants—mimicking the sentence fragments, brevity, and informal phrasing typical of the domain.
    \item \emph{SBERT (Sentence-Level Semantic Similarity):} SBERT computes embeddings for the entire sentence. The doubling of this metric (from 0.282 to 0.493) confirms that the model is capturing the \emph{holistic semantic intent} of the human response, ensuring that the generated ``noise'' preserves the underlying meaning of the parameter description.
\end{itemize}

Based on these results, we selected the checkpoint at \textbf{Epoch~2} for our experiments. While Epoch~3 achieves a marginally lower perplexity, we observe a degradation in semantic alignment, evidenced by the sharp drop in BERTScore (from 0.845 to 0.818). In contrast, Epoch~2 achieves the highest SBERT similarity ($0.493$) while maintaining a high BERTScore. This suggests that Epoch~2 offers the optimal trade-off: it preserves the highest semantic fidelity to human intent without overfitting to the specific lexical patterns of the training data. The result is a robust, scalable proxy that authentically simulates the ``last mile'' friction of inventory management.


    \section{Evaluation}\label{sec:eval}

In this section, we transition from theoretical design to empirical validation. Our evaluation is designed to measure the extent to which our hybrid agentic framework bridges the gap between \emph{descriptive reality} (ambiguous human inputs) and \emph{normative optimality} (mathematical best practices). Specifically, we aim to quantify the efficiency loss incurred when relying on the approximate reasoning of general-purpose LLMs rather than explicit stochastic optimization. By carefully isolating sources of error through controlled baselines, we seek to understand not only \emph{if} the system works, but also \emph{why} standard foundation models fail in this domain.

We address three primary research questions:

\begin{enumerate} 
\item \emph{Efficacy:} Does our hybrid architecture, by decoupling semantic understanding from logical optimization, outperform advanced commercial models (e.g., \emph{GPT-4o}) on complex stochastic inventory control tasks?

\item \emph{Structural Robustness:} Are the performance gains stable across varying degrees of environmental complexity? Under what conditions does our agentic framework achieve the largest marginal improvements?

\item \emph{Limits of Prompt-Based Reasoning:} Can the reasoning failures of general-purpose LLMs be remedied simply by providing perfect, ground-truth problem specifications? Or does stochastic optimization mark a fundamental \emph{computational boundary} that probabilistic language models cannot reliably cross without additional algorithmic and architectural support?

\end{enumerate}

\subsection{Evaluation Pipeline: An \textit{In Silico} Laboratory}

To empirically validate the Hybrid Agentic Framework proposed in \Cref{sec:framework}, we construct a controlled \textit{in silico} experiment using the Human Imitator developed in \Cref{sec:imitator}. This experimental design allows us to stress-test the system under realistic conversational friction, including ambiguity, inconsistency, and non-technical phrasing, while avoiding the logistical and financial constraints of large-scale human-subject trials.

Notably, this evaluation protocol generalizes beyond inventory management. By employing a high-fidelity synthetic agent to simulate the human-in-the-loop, our approach provides a scalable and reproducible testbed for evaluating a wide range of interactive LLM tasks where human variability and latent business knowledge are critical factors.

\subsubsection{Experimental Protocol.}\label{sec:protocol}
The evaluation protocol for each trial follows a three-stage lifecycle, as illustrated in \Cref{fig:evaluation}:

\begin{figure}[!t]
    \centering
    \includegraphics[width=0.75\linewidth]{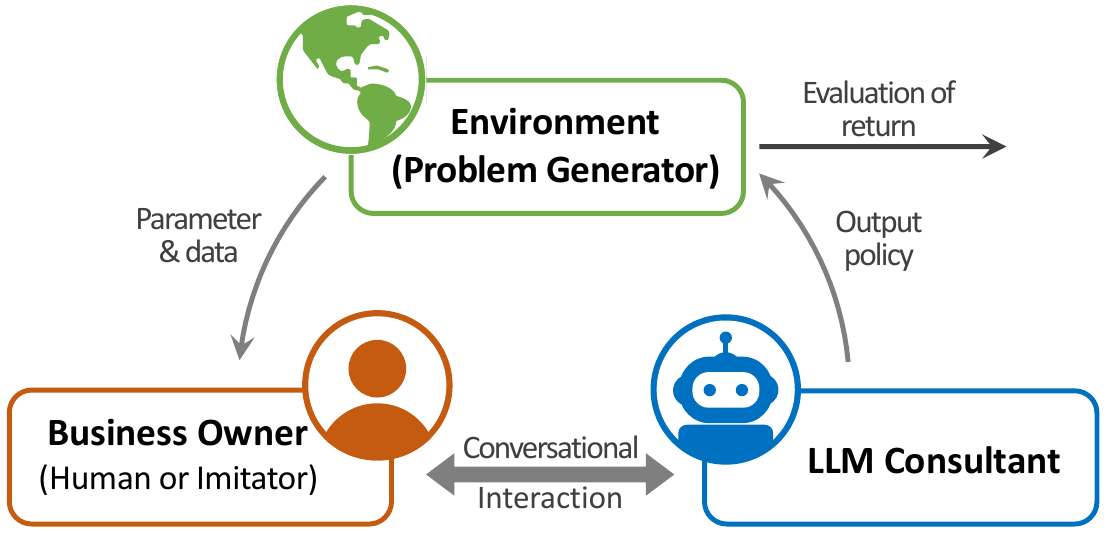} 
    \caption{\textbf{Evaluation Protocol.} The workflow moves from latent ground-truth generation \\ to conversational extraction, culminating in objective policy assessment.}
    \label{fig:evaluation}
\end{figure}

\begin{description}
    \item[1. Initialization (Ground Truth Generation):]
    The process begins with a {Problem Generator Agent}, implemented using \emph{GPT-4o}, which procedurally generates unique inventory management instances. Each instance comprises a high-level semantic context (e.g., ``Managing imported ingredient inventory for a premium restaurant'') and a precise set of \textit{ground-truth parameters} \footnote{When generating parameters, we follow standard parameterizations in the inventory management literature. For example, the inventory holding cost is set to 30\% of the item cost; the standard deviation of demand is approximately one-third of the mean; and the \emph{critical ratio}, defined as \texttt{penalty\_cost} / (\texttt{penalty\_cost} + \texttt{holding\_cost}), typically lies between 0.8 and 0.9.} (\Cref{tab:param-definitions}). These parameters constitute the ``latent reality'' of the business, representing details that a manager might intuitively grasp or access via records, yet struggles to articulate formally due to partial awareness or bounded rationality. Crucially, these ground-truth values provide the definitive benchmark against which all policies subsequently generated by the systems are evaluated.

    \begin{table}[!ht]
    \centering
    \renewcommand{\arraystretch}{1.2}
    \begin{tabular}{
        >{\raggedright\arraybackslash}p{0.26\textwidth}
        >{\raggedright\arraybackslash}p{0.705\textwidth}
    }
        \toprule
        \textbf{Parameter} & \textbf{Description} \\
        \midrule
        \texttt{time\_horizon} &
        The planning period for inventory decisions (default unit: days). \\

        \texttt{demand\_type} &
        The nature of customer demand, categorized as either \textit{deterministic} or \textit{random}. \\

        \texttt{demand\_distribution} &
        If deterministic, a constant value; if random, a statistical distribution (e.g., Normal($\mu=30, \sigma=10$)). \\
        & {\footnotesize (\emph{While our experiments primarily utilize Normal or Poisson distributions for simplicity, the framework supports a ``Custom'' setting that allows the system to infer distributions directly from user-uploaded historical data.})} \\

        \texttt{perishability} &
        A boolean value indicating if the items expire. \\

        \texttt{state\_transition\_model} &
        The outcome of a stockout, modeled as either a \textit{lost sale} or a \textit{backlog}. \\

        \texttt{holding\_cost} &
        The cost to store one unit of inventory for a specific time period. \\

        \texttt{penalty\_cost} &
        The cost incurred for each unit of unmet demand. \\

        \texttt{setup\_cost} &
        The fixed cost associated with placing an order. \\

        \texttt{lead\_time} &
        The time delay between placing an order and receiving it. \\

        \texttt{max\_inventory} &
        The maximum allowable inventory level. \\

        \texttt{max\_order} &
        The maximum quantity that can be ordered at one time. \\

        \texttt{risk\_tolerance} &
        An integer $\lambda$ in $\{ -10,-9, -8, \ldots, 9, 10\}$, where a lower value indicates higher risk aversion. \\
        \bottomrule
    \end{tabular}
    \caption{\textbf{Inventory Environment Parameters.} These variables define \\ the ground truth against which all policies are evaluated.}
    \label{tab:param-definitions}
\end{table}

    \item[2. Interaction:]
    We initialize the Human Imitator with the latent reality provided by the generator. It then engages a target system (selected from the treatment groups in \Cref{sec:benchmarks}) in a natural language dialogue. Crucially, the Imitator simulates the opacity of real-world consulting: rather than transmitting structured parameters directly, it describes the business situation conversationally, revealing information only through narrative descriptions and responses to follow-up inquiries, while maintaining realistic levels of ambiguity.
    
    \item[3. Assessment (Policy Evaluation):]
    Once the system proposes a policy, we evaluate its performance against the original \textit{ground-truth} parameters (distinct from the parameters the system may have inferred). We estimate the expected cost and standard deviation via Monte Carlo simulation.
    To accommodate different output formats, we employ a unified evaluation protocol: 
    \begin{itemize}[label=-]
        \item \emph{Automated Evaluation:} For standard mathematical policies (specifically Deep RL, $(s, S)$, or $(R, Q)$\footnote{The $(R, Q)$ policy is an inventory control strategy where a fixed quantity $Q$ is ordered whenever the inventory position drops to or below the reorder point $R$.}), our system automatically parses the parameters and executes the simulation.
        \item \emph{Manual Translation:} For policies proposed by baseline methods that do not fit these categories, we manually translate the described logic into executable code to ensure they are evaluated fairly.
    \end{itemize}
    Finally, we compute the optimality gap between the proposed policy and the theoretical benchmark.
\end{description}

This procedure is repeated across multiple trials to generate diverse inventory problem instances. Given that optimal costs vary significantly in magnitude across different scenarios, we report the relative cost difference between algorithms.

\subsubsection{Experimental Conditions (Baselines).}\label{sec:benchmarks}
\begin{figure}[!t]
    \centering
    \includegraphics[width=0.7\linewidth]{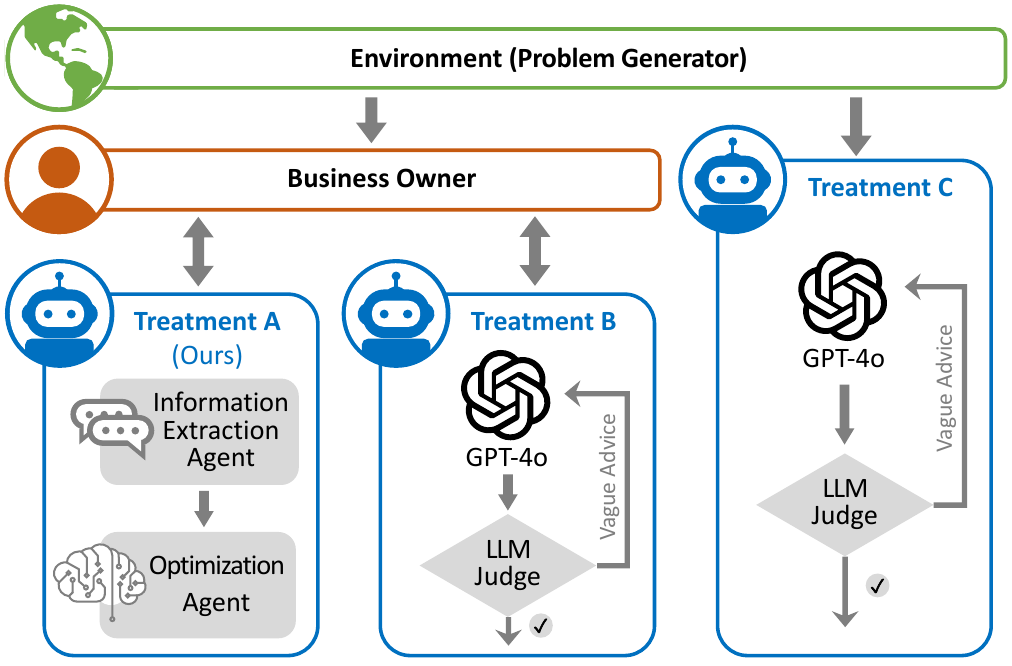} 
    \caption{\textbf{Overview of the comparative experimental framework.}}
    \label{fig:treatment}
\end{figure}

To isolate the sources of performance gain, we compare our proposed framework (in \Cref{sec:framework}) against two distinct \emph{GPT-4o} baselines. This comparative design allows us to decompose decision error into \textit{epistemic error} (translation failure) and \textit{computational error} (optimization failure).
\Cref{fig:treatment} provides a schematic overview of this experimental architecture, mapping the information flow from the latent ground truth (Environment) to the final mathematical policy (output) for each treatment.

\begin{description}
    \item[\textbf{Treatment A:} Interactive Hybrid Agent (Ours).] 
    This is the full proposed architecture in \Cref{sec:framework}. The system must navigate the conversation with the Human Imitator to extract parameters into a structured format (\Cref{tab:initial_params}) and then employ its internal Optimization Agent to compute a normative policy, using either a grid search for optimal $(s, S)$ thresholds or Deep Reinforcement Learning to handle complex, high-dimensional state spaces.

    \item[\textbf{Treatment B:} \textit{GPT-4o} (Interactive).] This baseline represents the standard ``End-to-End'' LLM approach. To ensure a rigorous and fair comparison, the Human Imitator initiates the conversation with the identical starting context used for Treatment A, the Hybrid Agentic System. Operating without access to external algorithmic solvers, the model must simultaneously navigate the dialogue context, infer the underlying operational parameters, and derive a policy based solely on its internal probabilistic weights. This treatment specifically tests the capacity of general-purpose models to overcome epistemic barriers and perform complex stochastic reasoning within a purely natural language interface.

    \item[\textbf{Treatment C:} \textit{GPT-4o} (Parameter Input).]
    This non-interactive baseline assesses the model's intrinsic inventory optimization capabilities under conditions of \textit{perfect information}. To bypass the ambiguity associated with conversational extraction, we directly feed the \textit{ground-truth parameter file}, a structured representation of the latent reality, to \emph{GPT-4o}, rather than relying on narrative extraction. The model is then explicitly prompted to derive an optimal policy. This counterfactual setup isolates computational reasoning from linguistic interpretation, allowing us to investigate whether the performance bottleneck in interactive settings stems from dialogue opacity or the model's fundamental algorithmic limitations.
\end{description}
\vspace{.5em}

\paragraph{Control Measures and Prompting.}
To ensure fairness and consistency, both \emph{GPT-4o} baselines (Treatments B \& C) are subject to strict output controls. Specifically, they are prompted with the following instruction to ensure the generation of \textit{evaluable policies} that are easily translatable into mathematical formulas, rather than vague qualitative advice:
\begin{tcolorbox}
    \scriptsize   
The final recommendation must have clear numbers, specify actions for all possible scenarios, and be easily translatable into a mathematical formula.

For example, ``When inventory falls below 20 units, order 80 units'' is a strategy.

``I recommend an (s, S) policy: when the sum of on-hand and on-order inventory drops below 35 units, order up to a total of 150 units'' is also a clear strategy.

``We should consider optimizing inventory'' or ``You need a better plan'' are not strategies.
\end{tcolorbox}
Without this instruction, \emph{GPT-4o} often generates underspecified recommendations—such as ``you need to order more when inventory is below 5 units''—which cannot be explicitly deployed or analyzed in simulation.

To operationalize these constraints, we further employ an auxiliary \textit{LLM Judge} (built on \mbox{\emph{GPT-4o}}). This judge monitors the response stream and acts as a termination mechanism: it validates whether the output contains a well-defined numerical policy and, upon confirmation, halts the generation to archive both the conversation log and the extracted policy for simulation.


    \subsection{Experimental Results and Findings}

{We report our findings based on a rigorous evaluation across $70$ randomly generated inventory scenarios. To ensure a standardized comparison, we evaluate performance based on expected cost minimization ($\mathds{E}\big[\mathrm{Cost}(\pi)\big]$), assuming a risk-neutral decision-maker (i.e. $\lambda = 10$ in \cref{eqn:objective}).}

{
\Cref{tab:cost_reduction_detailed} summarizes the comparative performance. The results provide strong empirical support for our architectural thesis: separating semantic understanding from logical optimization significantly outperforms end-to-end LLM reasoning.}

\begin{table}[!ht]
    \centering
    \small
    \begin{tabular}{l c c c}
        \toprule
        \textbf{Benchmark \& Scenario} & \textbf{Count ($n$)} & \textbf{\makecell{Avg. Cost Red.\\(Mean \& Std. Dev.)}} & \textbf{\makecell{Win Rate\\(\%)}} \\
        \midrule
        \multicolumn{4}{l}{\textit{\textbf{Baseline 1: vs. GPT-4o (Interactive)}}} \\
        \hspace{1em}Overall Performance & 70 & \textbf{32.1\%} $\pm$ 17.8\% & 97.1\% \\
        \addlinespace[0.5em]
        \hspace{1em}\textit{By Demand Distribution (Lead Time = 7 days)} & & & \\
        \hspace{2em}Normal Distribution & 14 & 34.2\% $\pm$ 15.0\%& 100\% \\
        \hspace{2em}Poisson Distribution & 11 & 33.8\% $\pm$ 19.7\%& 100\%\\
        \addlinespace[0.5em]
        \hspace{1em}\textit{By Lead Time (Demand = Poisson)} & & & \\
        \hspace{2em}Short Lead Time (1 day)  & 15 & 26.8\% $\pm$16.4\% & 93.3\% \\
        \hspace{2em}Long Lead Time (7 days) & 11 & 33.8\% $\pm$19.7\% & 100\% \\
        \midrule
        \multicolumn{4}{l}{\textit{\textbf{Baseline 2: vs. GPT-4o (Parameter Input)}}} \\
        \hspace{1em}Overall Performance & 70 & \textbf{33.4\%} $\pm$ 16.8\% & 100\% \\
        \bottomrule
    \end{tabular}
    \caption{\textbf{Performance Statistics of Hybrid Agentic Framework.} \\
    This table reports the cost reduction statistics (Mean and Standard Deviation) and \\ the win rate (percentage of instances where the Hybrid Agent outperformed the benchmark).}
    \label{tab:cost_reduction_detailed}
\end{table}

To contextualize the economic scale of these experiments, our simulation parameters mirror the operating realities of Small and Medium-sized Enterprises (SMEs). The holding cost in our testbed ranges from \$0.25 to \$4.00 per unit/period, with penalty costs for stockouts ranging significantly higher (averaging roughly \$5.00). Over a standard planning horizon of a few months, the cumulative cost for a single item typically runs into the thousands of dollars. While these unit economics may appear modest in isolation, they scale linearly across the hundreds of SKUs typical of a retail inventory. Consequently, a persistent efficiency gap of roughly 30\% translates into substantial margin erosion for a real-world business, validating the material significance of our findings.

{In the following subsections, we unpack the sources of this advantage in three steps. First, we quantify the aggregate efficiency gap—the ``Hallucination Tax''—in interactive settings (\Cref{sec:hallucination_tax}). Second, we identify the operational conditions that make the agent most valuable (\Cref{sec:driver}). Finally, we isolate the root cause of the baseline's failure by distinguishing between errors of understanding (bad inputs) and errors of reasoning (bad logic) (\Cref{sec:cognitive_ceiling}).}

\subsubsection{The Hallucination Tax (\emph{Treatment A} vs. \emph{B}).}
\label{sec:hallucination_tax}

Our comparison begins with the aggregate performance gap between our Hybrid Agentic Framework (\emph{Treatment A}) and the \emph{GPT-4o} Interactive baseline (\emph{Treatment B}). As detailed in \Cref{tab:cost_reduction_detailed}, the hybrid approach yields a substantial $\mathbf{32.1\%}$ cost reduction (standard deviation $17.8\%$) and achieves a $97.1\%$ win rate across the $70$ test instances. In the rare exceptions ($2$ instances) where \emph{Treatment~B} performs better, the margins are negligible ($2\%$ and $6\%$, respectively), and in both cases, the baseline happens to converge to an $(s, S)$ heuristic effectively identical to the one identified by our optimizer.

We term this efficiency loss the \textbf{``Hallucination Tax.''} It represents the compound penalty incurred when a firm relies on a general-purpose language model as an end-to-end solver. This gap is not merely a reflection of random error, but a structural divergence in how the two systems approach problem-solving: one relies on probabilistic token generation, the other on rigorous mathematical optimization.

\vspace{.5em}

\paragraph{The Fluency Trap: Why Managers Get Fooled.}
A qualitative analysis of the conversation logs reveals a dangerous phenomenon we call the \emph{fluency trap}. The \emph{GPT-4o} baseline is highly eloquent; it correctly uses domain terminology and offers structurally plausible advice. However, this linguistic competence masks a fundamental inability to calibrate policy parameters to specific constraints.

A representative example is observed in the dialogue in \Cref{fig:example1_part2}. Despite explicitly acknowledging a maximum inventory capacity of $80$ units, the \emph{GPT-4o} model confidently recommends: \emph{``When your total inventory... drops to or below 89 units, place an order to bring the inventory up to 80 units.''} 
This recommendation is not merely suboptimal; it is physically impossible. 

For a manager, the model's confident tone creates a false sense of security. The ``Hallucination Tax'' quantifies the cost of this illusion: the difference between a decision that \emph{sounds} correct and one that \emph{is} feasible and optimal. For an SME retailer with net margins of $3$--$5\%$, relying on the ``plausible but suboptimal'' advice of a raw LLM could be the difference between profitability and insolvency.

\subsubsection{Drivers of Performance: When is the Agent Most Valuable?}
\label{sec:driver}

To identify the structural conditions where our framework offers the greatest marginal gains over \emph{GPT-4o} (Interactive), we decompose the results along three dimensions. Our analysis reveals that the value of the Hybrid Framework is not uniform; rather, it scales non-linearly with the complexity and economic consequences of the control problem. \vspace{.5em}

\paragraph{Distribution Agnosticism (Robustness against Tail Risk).}
We examine the system's resilience to demand uncertainty by testing across distinct distribution profiles: Normal distributions (representing high-volume, symmetric demand typical of staples) and Poisson distributions (representing low-volume, right-skewed demand typical of erratic or slow-moving items). 
When controlling for lead time (7 days), the agent achieves a $34.2\%$ cost reduction on Normal instances and a comparable $33.8\%$ reduction on Poisson instances. The consistency of these results indicates that the gains from our solver-backed approach are largely agnostic to the underlying demand distribution. 

In inventory control, optimal policies are often determined not by the \emph{average} demand, but by the probability of extreme events in the tail. LLMs, relying on linguistic patterns, tend to exhibit ``bias towards the mean,'' often underestimating the risk of stockouts in skewed (Poisson) scenarios or over-buffering in symmetric (Normal) ones. In contrast, our solver-backed approach explicitly integrates over the full probability density function. This allows the system to remain \emph{distribution agnostic}—delivering precise control for both steady staples and volatile long-tail items. \vspace{.5em}

\paragraph{The Value of Temporal Depth (Impact of Lead Time).}
We evaluate the system's performance across varying supply chain latencies: {Short Lead Times} ($L=1$ day, representing rapid replenishment) and {Long Lead Times} ($L=7$ days, representing cross-regional logistics with delayed fulfillment).
On Poisson-demand instances, the performance gap widens significantly with latency. The cost reduction starts at $26.8\%$ for short lead times and expands to $33.8\%$ for long lead times. 

This divergence stems from the challenge of managing \emph{pipeline inventory} (orders placed but not yet received). As lead times grow, a significant portion of stock is ``hidden'' in transit. Basic LLMs and simple heuristics tend to suffer from \emph{myopia}: reacting to immediate on-hand levels while ignoring incoming shipments. In contrast, our Optimization Agent explicitly tracks these in-transit orders as part of the system state. The widening gap indicates that while simple heuristics may suffice when replenishment is nearly immediate, they fail to master the complex intertemporal trade-offs required in high-latency supply chains, where our rigorous solver provides a clearer advantage.  \vspace{.5em}

\paragraph{The Complexity Premium (Impact of Economic Stakes and Action Space).}

We further isolate the parameters driving the largest marginal gains by comparing the structural characteristics of the top and bottom quartiles ($25\%$) of performance improvement. As summarized in \Cref{tab:quartile_analysis}, the analysis confirms a \emph{complexity premium}: the agent's comparative advantage is closely tied to the financial stakes and the richness of the action space.

\begin{table}[!ht]
    \centering
    \begin{tabular}{l c c c}
        \toprule
        \textbf{Structural Parameter} & \textbf{\makecell{Top Quartile\\(High Gain)}} & \textbf{\makecell{Bottom Quartile\\(Low Gain)}} & \textbf{\makecell{Significance\\($p$-value)}} \\
        \midrule
        \multicolumn{4}{l}{\textit{\textbf{Economic Stakes}}} \\
        \hspace{1em}Holding Cost & \$1.19 & \$0.71 & $< 0.01$ \\
        \hspace{1em}Penalty Cost & \$6.08 & \$3.84 & $< 0.01$ \\
        \addlinespace[0.5em]
        \multicolumn{4}{l}{\textit{\textbf{Action Space Complexity}}} \\
        \hspace{1em}Max Order Limit & 32.2 & 26.4 & $< 0.02$ \\
        \addlinespace[0.5em]
        \multicolumn{4}{l}{\textit{\textbf{Control Variables (Non-Significant)}}} \\
        \hspace{1em}Set-up Cost & \$3.08 & \$3.06 & $\geq 0.05$ \\
        \hspace{1em}Inventory Capacity & 73.3 & 72.2 & $\geq 0.05$ \\
        \bottomrule
    \end{tabular}
    \caption{\textbf{The Complexity Premium: Structural Characteristics of Top vs. Bottom Quartiles.} \\
    This table compares the average parameter values for instances where the Hybrid Agent achieved the highest cost reductions versus the lowest. While economic stakes and action space show significant divergence, other structural constraints (set-up cost, capacity) do not.}
    \label{tab:quartile_analysis}
\end{table}

\begin{itemize}
    \item \emph{High Economic Stakes Amplify the Cost of Error.} 
    The most significant differentiator is the unit cost structure. Instances in the top quartile exhibit holding costs that are roughly $68\%$ higher and penalty costs that are $58\%$ higher than those in the bottom quartile. This indicates that simple heuristics or vanilla LLMs may perform adequately when items are cheap and the ``forgiveness'' for error is high. However, when the trade-off between holding inventory and losing sales is financially consequential, the precision of a formal solver becomes indispensable.
    
    \item \emph{Broader Action Spaces Reward Optimization.} 
    The top quartile is also characterized by a significantly looser maximum order limit ($32.2$ vs. $26.4$). A tighter limit artificially constrains the solution space, forcing all policies—optimal or heuristic—to converge. Conversely, a broader limit expands the decision boundaries. Our Agentic Framework exploits this flexibility to discover sophisticated replenishment strategies, whereas the baseline tends to fall back on conservative, static heuristics that fail to use the full range of valid actions.
\end{itemize}

It is worth noting that other structural constraints, such as set-up cost and inventory capacity, showed no statistically significant divergence. Taken together, these findings suggest that the Hybrid Agentic Framework is especially beneficial in high-cost, high-flexibility settings, where precision in policy optimization matters most and naive heuristics leave substantial value on the table.

\subsubsection{Isolating the Source of Error: The Failure of Perfect Information (\mbox{\emph{Treatment A}} vs. \emph{C}).}
\label{sec:cognitive_ceiling}

Given the significant efficiency loss observed in the interactive setting, a natural question arises: \emph{Is this failure driven by linguistic ambiguity or computational limitation?} One might hypothesize that the LLM underperforms primarily because the messiness of natural language prevents it from extracting the correct problem parameters.
\emph{Treatment C} tests this hypothesis by removing the communicative burden entirely. We feed the ``ground-truth parameter file'' directly to \emph{GPT-4o}, granting it a complete, noise-free view of the inventory environment. If the bottleneck were solely informational, performance should converge toward the optimal benchmark.
\vspace{-1em}

\paragraph{The Persistence of the Heuristic Tax.}The empirical results decisively reject this hypothesis. As shown in \Cref{tab:cost_reduction_detailed}, providing perfect information yields no statistical improvement over the noisy interactive baseline ($p$-value $= 0.66$). The system hits a hard \emph{Cognitive Ceiling}.

Revisiting the representative instance from \Cref{sec:example1} clarifies exactly what improves—and what remains broken—under perfect information:
\begin{itemize}
\item \emph{In Treatment B (Interactive),} the noise of conversation leads to logical incoherence. The model hallucinates a reorder point ($89$) that violates the capacity constraint ($80$). This is an \emph{Interpretive Error}.
\item \emph{In Treatment C (Perfect Info),} this specific hallucination disappears. With clear access to the parameters, the model respects the capacity limit. However, it still recommends a suboptimal policy (e.g., $(40, 65)$) based on a heuristic estimation of the mean, failing to optimize for the tail risk of stockouts.
\end{itemize}
This reveals that while better data can cure logical hallucinations, it cannot cure the fundamental deficit in stochastic reasoning.
\vspace{.5em}

\paragraph{The Indispensability of Operations Research.}This finding brings us to a critical architectural conclusion regarding the role of General Purpose AI. One might argue that future iterations of foundation models will eventually close this gap without the need for external solvers. Our results suggest otherwise.
Relying on a generalist model to perform stochastic optimization is inherently inefficient and unreliable. To replicate the precision of a dedicated OR solver, a Large Language Model must generate extensive chains of thought, consuming vast amounts of tokens and computational resources to approximate what a standard algorithm (e.g., dynamic programming) can solve in milliseconds. Furthermore, probabilistic token generation remains fundamentally prone to stochastic drift—it can produce numbers that \emph{look} plausible but are mathematically groundless.
Therefore, the ``Last Mile'' of decision support cannot be bridged by scaling model size alone. It requires a division of labor: using the LLM for what it does best (semantic interpretation) and the OR solver for what it does best (deterministic optimization). This hybrid architecture is not merely an engineering patch but a necessary epistemological bridge between descriptive language and normative optimality.

\subsection{Qualitative Analysis on a Representative Instance.}\label{sec:example1}

We conclude this section by presenting conversation logs from \emph{Treatments A--C} on a representative instance. Example 1 demonstrates a scenario where the user approaches the system with a broad background description of a bike repair shop. This behavior is typical of the interactions observed in our experiments.

\begin{figure}[!t]
    \centering
    \includegraphics[width=\textwidth, keepaspectratio]{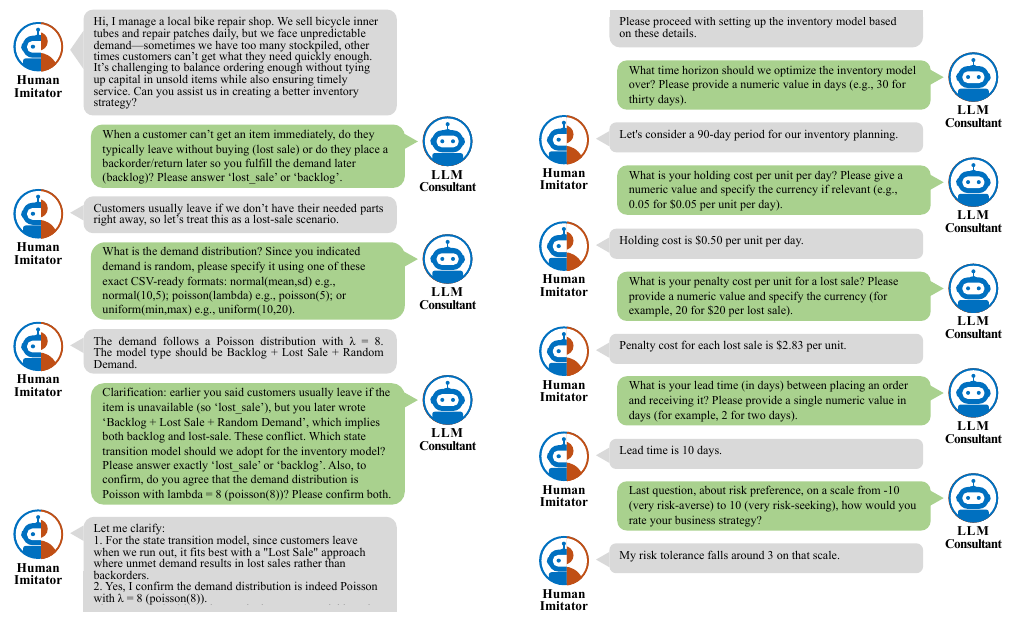}
    \caption{\textbf{Example 1 - Treatment A.} Complete conversation logs (Interactive Agent).}
    \label{fig:example1_part1}
\end{figure}

\begin{figure}[!t]
    \centering
    \includegraphics[width=\textwidth, keepaspectratio]{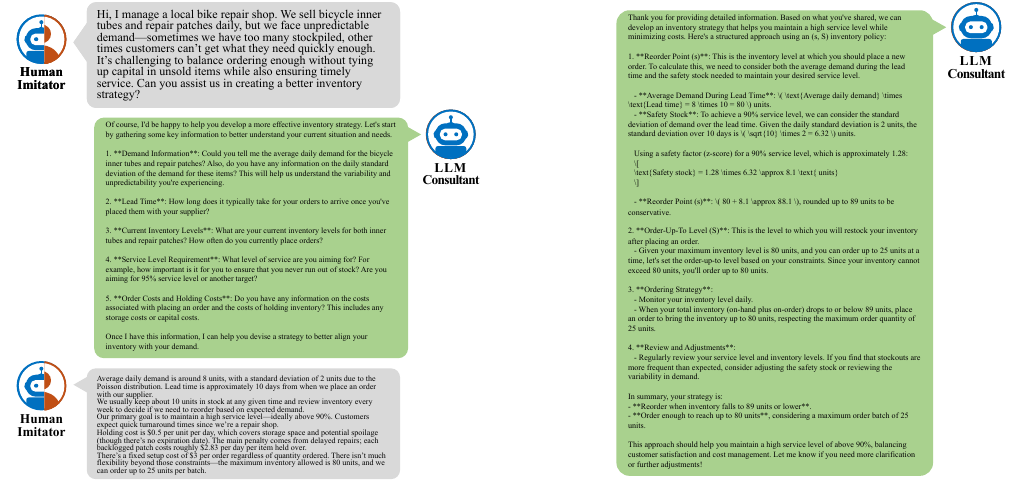}
    \caption{\textbf{Example 1 - Treatment B.} Complete conversation logs (GPT-4o Interactive).}
    \label{fig:example1_part2}
\end{figure}

\begin{figure}[!t]
    \centering
    \includegraphics[width=0.7\textwidth, keepaspectratio]{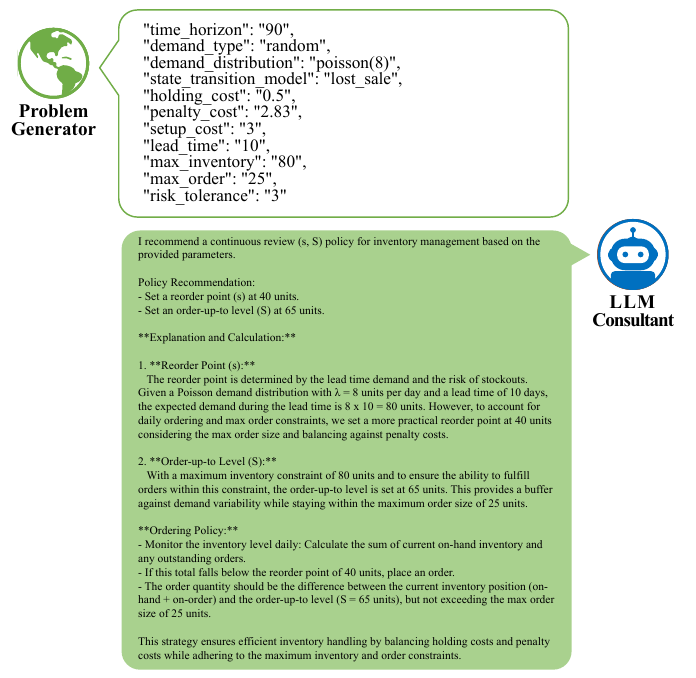}
    \caption{\textbf{Example 1 - Treatment C.} Complete conversation logs (GPT-4o Parameter Input).}
    \label{fig:example1_full}
\end{figure}

Our \textit{Agentic Framework (Treatment A)} (\Cref{fig:example1_part1}) follows a rigorous slot-filling and verification process. Notably, the agent detects a logical inconsistency when the user initially specifies a ``lost-sale'' model but later requests a ``Backlog + Lost Sale'' model. The agent explicitly flags this conflict (``These conflict'') and prompts the user to select a single valid transition model. After approximately nine rounds of dialogue to resolve such ambiguities and gather all necessary parameters (e.g., time horizon, costs, capacities), the agent forwards a verified parameter set to the Optimization Agent.

In contrast, the \textit{GPT-4o Interactive (Treatment B)} baseline (\Cref{fig:example1_part2}) collects relatively less detailed information, typically asking only for summary statistics (e.g., mean and standard deviation) instead of the full demand distribution required for rigorous modeling. Moreover, it relies on internal heuristics to generate a policy, often leading to hallucinations or feasibility violations. Crucially, in this instance, the baseline recommends an $(s, S)$ policy with a reorder point of 89 units, despite previously acknowledging a maximum inventory capacity of only 80 units. This recommendation is physically impossible to implement. The \textit{GPT-4o Parameter Input (Treatment C)} baseline (\Cref{fig:example1_full}) generates yet another distinct policy ($(s, S)$ with $s=40, S=65$), highlighting the inconsistency and lack of robustness in direct LLM reasoning compared to our solver-backed approach.

Additional conversation logs are provided in \Cref{sec:sample_conversation}. In most instances, the \textit{GPT-4o} baselines rely on simple heuristics to generate standard policies, such as $(s, S)$ or $(R, Q)$. Even in instances where the $(s, S)$ policy structure is preferred, our system applies grid search to mathematically optimize the $s$ and $S$ parameters, whereas \emph{GPT-4o} merely approximates them using basic heuristics.


\section{Managerial Implications and Industry Perspectives}
\label{sec:implications}

Our findings extend beyond algorithmic validation to inform the design of decision support in next-generation Enterprise Resource Planning (ERP) systems. While conversational assistants can materially improve accessibility to data and routine workflows, our results suggest a practical limitation: when systems move from \emph{insight generation} to \emph{operational control}, reliability hinges less on linguistic fluency and more on explicit optimization, constraint enforcement, and verifiable execution. This motivates an architectural pattern in which the natural-language interface is separated from the decision engine, rather than treating a general-purpose LLM as an end-to-end controller.

\subsubsection*{Beyond the ``Chatbot'' Paradigm: Where Generalist AI Struggles.}
Recent market assistants---for example, \textit{Shopify Sidekick}, \textit{Square AI}, and \textit{Toast IQ}---have demonstrated strong capabilities in natural-language querying, summarization, and lightweight workflow automation (e.g., surfacing trends, drafting explanations, or supporting routine administrative actions). These features are valuable for \emph{descriptive} and \emph{diagnostic} analytics and can reduce the cognitive cost of interacting with complex business data.

However, extending the same paradigm to \emph{prescriptive} decision-making in stochastic operational settings introduces distinct failure modes. 
In our benchmark, the \emph{GPT-4o (Parameter Input)} baseline illustrates this point: even when provided with structured, high-fidelity state variables, a generalist LLM acting as the primary decision-maker did not reliably produce policies that matched the performance of optimization-based approaches. Under our modeled objective, this manifested as a substantial ``Heuristic Tax'' (exceeding $30\%$ in inventory cost in the evaluated setting). 
Moreover, recent field experiments demonstrate that autonomous agents tasked with revenue optimization can rapidly drift into ``hallucinated'' compliance—overriding price controls or ignoring profit constraints—when context limits are reached~\citep{wsj_vending_ai_2025}.
Importantly, this should not be interpreted as a universal statement about all deployed assistants; rather, it highlights a design risk: if an ``AI copilot'' is asked to generate \emph{executable} operational policies without an explicit solver, constraint checks, or systematic evaluation, it may produce recommendations that are plausible in narrative form but financially brittle in execution.

The managerial implication is therefore not to avoid LLM-based interfaces, but to scope them appropriately: use generalist models to \emph{elicit intent, clarify constraints, and explain trade-offs}, while delegating the computation of actions (and their validation) to specialized routines with auditable guarantees.

\subsubsection*{Lowering the Barrier to Operational Rigor.}
Advanced inventory policies (e.g., dynamic $(s, S)$ rules derived from dynamic programming or approximate stochastic control) have long been available in the academic and industrial toolkit. In practice, however, deploying these methods at scale has been disproportionately easier for firms with strong data infrastructure, analytics talent, and process discipline. Small and medium-sized enterprises (SMEs) often operate under constraints---limited analyst capacity, inconsistent master data, and tight operational bandwidth---that make it difficult to translate formal models into day-to-day decisions. As a result, many SMEs rely on stable heuristics (e.g., fixed reorder points or weekly manual adjustments), not because optimization is conceptually inaccessible, but because the \emph{implementation and maintenance costs} are nontrivial.

Our hybrid agentic framework targets this deployment gap. By acting as a semantic bridge, the system allows decision-makers to express business intent in natural language (service targets, ordering constraints, budget limits, and operational preferences) while executing the resulting decision problem with a solver-based backend. In this sense, the framework can \emph{productize} operational rigor: it reduces the expertise required to \emph{use} advanced policies, even though the policies themselves remain grounded in explicit formulations and validated computation.

The economic implications should be interpreted with appropriate caution and context. In our benchmark environment, the proposed approach achieved a 32.1\% reduction in the modeled inventory cost relative to heuristic baselines. If improvements of comparable direction translate to practice, they may materially affect cash flow and working-capital utilization, especially in sectors with thin margins and limited buffer capacity. That said, realized impact will vary with demand volatility, lead-time uncertainty, the baseline inventory-to-sales ratio, and execution frictions (data quality, supplier constraints, and organizational adherence). Managers should therefore treat performance gains as \emph{evaluable hypotheses} that require backtesting and staged rollouts rather than as guaranteed outcomes.

\subsubsection*{The Future of Intelligent ERP: A Three-Stage Design Pattern.}
We frame the evolution of management software as a design pattern with three stages:
\begin{itemize}
    \item \textbf{Type I (Descriptive):} Dashboards and reports that visualize data and leave decision logic entirely to human operators.
    \item \textbf{Type II (LLM-Augmented):} Systems that add conversational interfaces and agentic features to summarize data, draft recommendations, and trigger limited actions. Decision logic in these systems is often heuristic, partially opaque, or weakly constrained, which can be problematic for high-stakes operational control.
    \item \textbf{Type III (Orchestrating):} Systems in which the LLM is a \emph{front-end orchestrator} rather than the decision engine.
\end{itemize}

In a Type III system, the interface and the engine are intentionally decoupled. The LLM is used for (i) intent capture and constraint elicitation, (ii) translation into structured problem specifications, and (iii) explanation and what-if communication. Execution is delegated to specialized optimization and simulation routines, with guardrails such as constraint validation, audit logs, approval workflows for high-impact actions, and fallback policies when assumptions drift. This architecture does not eliminate uncertainty or model risk; instead, it makes the locus of decision-making explicit and testable.

We therefore view modular orchestration as a robust pathway toward ``prescriptive ERP'' in domains where decisions are frequent, costs are nonlinear, and uncertainty is material. The central managerial takeaway is to invest in systems that can \emph{communicate naturally} while remaining \emph{solver-validated under a specified formulation}, and to operationalize them with monitoring and governance rather than treating language fluency as a proxy for decision quality.

\section{Conclusion and Future Directions}
\label{sec:conclusion}

A central tension in management science lies between \emph{normative optimality}—the theoretical best prescribed by models—and \emph{descriptive reality}—the ambiguous, messy context in which practitioners operate. This paper proposes a Hybrid Agentic Framework to bridge this gap. By combining the semantic flexibility of Large Language Models with the rigorous precision of Operations Research solvers, we demonstrate that it is possible to democratize advanced inventory control without sacrificing mathematical fidelity.
Our empirical evaluation, conducted on a novel \emph{Human Imitator} testbed, confirms that this division of labor is superior to end-to-end LLM reasoning.

Our framework is designed as a modular foundation, inviting extension across several dimensions:

\begin{itemize}\item \textbf{Ecological Validation:} While our \emph{Human Imitator} provides a scalable proxy for bounded rationality, future work should validate the system against real-world human subjects. This involves refining the information extraction pipeline to handle the nuances, interruptions, and non-linear logic typical of actual manager-consultant dialogues.\item \textbf{Algorithmic and Agentic Expansion:} The current \emph{Optimization Agent} is agnostic to the underlying solver; future iterations could incorporate more advanced methodologies. Furthermore, we envision richer interaction patterns, such as a ``Critic Agent'' that detects inconsistencies between user goals and solver outputs, creating a verification loop before recommendations are presented.

\item \textbf{Scope Generalization:} Finally, this architecture can be extended beyond inventory management. By swapping the underlying solver, the framework could support other high-stakes decision domains—such as dynamic pricing, workforce scheduling, or logistics routing—where the synergy of natural language interfaces and rigorous optimization is critical.
\end{itemize}

\section*{Acknowledge}

    Duan was supported in part by NSF DMS-2413812 and the LSE--NYU Research Seed Fund.
    
    The authors thank Vishal Gaur, Vrinda Kadiyali, Kaizheng Wang, Karen Jiaxi Wang, Linwei Xin, and Jiawei Zhang for helpful discussions and feedback.


\bibliographystyle{abbrvnat}
\bibliography{files/ref}

\clearpage

\OneAndAHalfSpacedXI

%
%
%

\begin{APPENDICES}
\section{Inventory Model Overview}\label{sec:InventoryModel}

We focus on the fundamental single-product inventory control problem to illustrate our general framework. However, our approach can be directly extended to the multi-problem or multi-echelon cases. To be specific, we 
consider a periodic-review inventory system for a single product over a finite horizon of $T$ periods with stochastic demand. In each period $t$, demand $D_t$ lies in $[0, \bar{D}]$ and is drawn independently from an unknown distribution $F(\cdot)$. The firm places an order $q_t$ at the beginning of period $t$, which arrives after a deterministic lead time $L \in \mathds{N}$. Let $h \ge 0$ denote the per-unit holding cost, $b \ge 0$ the per-unit lost-sales penalty, and $H$ the setup cost.
The sequence of events within each period $t$ is as follows:
\begin{enumerate}
    \item \textbf{State observation:} The firm observes on-hand inventory $I_t$ and the pipeline vector $(x_{1,t}, \ldots, x_{L,t})$, where $x_{i,t}$ is the order placed at $t-L+i-1$ for $i = 1, \ldots, L$. The system state is $(I_t, x_{1,t}, \ldots, x_{L,t})$.
    \item \textbf{On-hand inventory update:} The order due to arrive is received, and on-hand inventory updates to $I_t + x_{1,t}$.
    \item \textbf{Ordering decision:} The firm places $q_t$, which will arrive at the beginning of period $t+L$ (i.e., \mbox{$x_{L,t+1} = q_t$}).
    \item \textbf{Demand realization and fulfillment:} Demand $D_t$ realizes and is satisfied up to available inventory; unmet demand is lost and unobserved.
\end{enumerate}
If the system is a lost-sale system and the product is non-perishable, then
the state evolves according to
\[
I_{t+1} = \big(I_t + x_{1,t} - D_t\big)^+, 
\quad x_{i,t+1} = x_{i+1,t} \ \text{ for } 1 \le i \le L-1, 
\quad x_{L,t+1} = q_t.
\]
In contrast, if the system is a back-order system and the product is non-perishable, then the state evolves according to 
\[
I_{t+1} = I_t + x_{1,t} - D_t, 
\quad x_{i,t+1} = x_{i+1,t} \ \text{ for } 1 \le i \le L-1, 
\quad x_{L,t+1} = q_t,
\]
where we allow the on-hand inventory level to be negative which will be replenished by future orders. Finally, if the product is perishable and the system is lost-sale, then the state evolves according to
\[
I_{t+1} = 0, 
\quad x_{i,t+1} = x_{i+1,t} \ \text{ for } 1 \le i \le L-1, 
\quad x_{L,t+1} = q_t.
\]
If the product is perishable and the system is back-order, then the state evolves according to
\[
I_{t+1} = \min\left\{ I_t + x_{1,t} - D_t, 0  \right\}, 
\quad x_{i,t+1} = x_{i+1,t} \ \text{ for } 1 \le i \le L-1, 
\quad x_{L,t+1} = q_t.
\]
A policy $\policy$ specifies order quantities $q_1^\policy, \ldots, q_T^\policy$. A policy $\policy$ is \emph{feasible} if it is non-anticipative: for each $t$, $q_t^\policy$ depends only on past and current states $\{(I_\tau^\policy, x_{1,\tau}^\policy, \ldots, x_{L,\tau}^\policy): \tau \le t\}$ and realized demand information $\{D_{\tau}: \tau \le t\}$. The distribution $F(\cdot)$ is unknown and must be learned.
The period-$t$ cost under policy $\policy$ is
\[
C_t^\policy \;=\; H\cdot \mathds{1}_{\{q^\policy_t>0\}} + h \cdot \big(I_t + x_{1,t} - D_t\big)^+ \;+\; b \cdot \big(D_t - I_t - x_{1,t}\big)^+.
\]
The expected cumulative cost over $T$ periods is
\[
C^\policy(T, L) \;=\; \sum_{t=1}^T \mathds{E}[C_t^\policy]
\;=\; \sum_{t=1}^T \mathds{E}\!\left[H\cdot \mathds{1}_{\{q^\policy_t>0\}} + h \cdot \big(I_t + x_{1,t} - D_t\big)^+ + b \cdot \big(D_t - I_t - x_{1,t}\big)^+\right].
\]
We develop policies to minimize the long-run average cost given by $C^\policy(T, L)$.

\subsection{An Overview of Heuristic Policies}
We introduce several heuristic policies for the periodic-review, lead-time inventory problem described above, covering lost-sales and backorder systems as well as perishable and non-perishable products. 

\noindent\textbf{Order-up-to policy}.
A natural starting point is the \textit{order-up-to}, or \textit{base-stock policy}. In each period, the firm raises its inventory position---defined as the sum of on-hand inventory and the $L$ outstanding pipeline orders---to a target level $S$ by ordering $q_t = [S - \mathrm{IP}_t]^+$, where $\mathrm{IP}_t = I_t + \sum_{i=1}^L x_{i,t}$. Under non-perishable items with backorders, linear holding and backorder costs, and i.i.d.\ demand, base-stock policies are often optimal; the target $S$ can be set using the newsvendor fractile for the $L$-period protection demand, when the distribution is given, or by set using a bi-section method when the distribution is unknown and only historical samples are available. In lost-sales systems, base-stock policies remain widely used though optimality is more delicate; in practice, $S$ is tuned by minimizing an approximate expected holding and lost-sales cost over the protection period.

\noindent\textbf{Constant-order policy}. The \textit{constant-order policy} prescribes ordering the same quantity in every period, independent of the current state. Formally, fix a constant $q \ge 0$ and set $q_t \equiv q$ for all $t$. The constant $q$ can be chosen using historical demand information, e.g., $q \approx \mathds{E}[D]$, or tuned via optimization or simulation to balance holding and shortage costs under the given lead time $L$. This policy is particularly appealing in lost-sales systems with long lead times and high penalty $b/h$, where smoothing the pipeline can mitigate stockouts while avoiding excessive accumulation. Variants include \textit{capped constant-order} policies, where the order is truncated to respect capacity constraints, $q_t = \min\{\bar{Q}, q\}$, or combined with minimal state dependence through caps on the inventory position, $q_t = \min\{\bar{Q}, [S - \mathrm{IP}_t]^+\}$ with a fixed $S$.

\noindent\textbf{$(s, S)$ policy}.
We now introduce the \textit{$(s, S)$ policy}, a classical state-dependent rule that generalizes the base-stock policy and is particularly suitable when the setup cost $H$ is positive. The $(s, S)$ policy is characterized by two parameters with $0 \le s \le S$. In each period $t$, the firm observes its inventory position $\mathrm{IP}_t = I_t + \sum_{i=1}^L x_{i,t}$ and orders only if $\mathrm{IP}_t < s$; when an order is placed, it raises the position up to $S$. Formally, the order quantity is
\[
q_t^{(s,S)} \;=\; \big[\, S - \mathrm{IP}_t \,\big]^+ \cdot \mathds{1}\{\mathrm{IP}_t < s\}.
\]
Intuitively, $s$ serves as a reorder threshold that limits the frequency of orders, while $S$ controls the target protection level against lead-time demand. In non-perishable, backorder systems with linear holding and backorder costs plus a fixed setup cost, $(s, S)$ policies are often optimal; absent a fixed cost, the rule reduces to a base-stock policy with $s = S$. In lost-sales systems, $(s, S)$ remains a practical choice, where $S$ is tuned to the desired service level or the newsvendor quantile for the protection period and $s$ trades off order frequency and pipeline smoothing. Capacity constraints can be easily incorporated via truncation 
\[
q_t^{(s,S)} = \min\{\bar{Q}, \, [S - \mathrm{IP}_t]^+ \mathds{1}\{\mathrm{IP}_t < s\}\}.
\]
We refer interested readers to \cite{simchi2005logic} for the introduction of other heuristic policies.
\subsection{General Policies via MDP Formulation}
We now cast the problem as a Markov decision process. For non-perishable items with lead time $L$, the system state at the beginning of period $t$ is $s_t = (I_t, x_{1,t}, \ldots, x_{L,t})$, where $I_t$ denotes on-hand inventory (which may be negative in backorder systems) and $x_{i,t}$ denotes the pipeline order scheduled to arrive in $i$ periods. The action is the nonnegative order quantity $q_t$ chosen at the ordering epoch. Exogenous uncertainty enters through the demand $D_t$, which is i.i.d.\ over time and distributed according to an unknown distribution $F$ supported on $[0, \bar{D}]$. The state transition follows the operational sequence: the oldest pipeline order $x_{1,t}$ arrives before demand, the pipeline shifts forward with $x_{i,t+1} = x_{i+1,t}$ for $i = 1, \ldots, L-1$ and $x_{L,t+1} = q_t$, and the on-hand inventory updates according to the fulfillment convention. 

In the non-perishable lost-sales model, $I_{t+1} = (I_t + x_{1,t} - D_t)^+$, whereas in the non-perishable backorder model, $I_{t+1} = I_t + x_{1,t} - D_t$. For perishable products with one-period shelf life under lost sales, any leftover inventory perishes at the end of the period and $I_{t+1} = 0$; for perishable products with backorders, we set $I_{t+1} = \min\{I_t + x_{1,t} - D_t, 0\}$ to reflect that unmet demand carries forward while excess perishables do not.

The period cost $c(s_t, q_t, D_t)$ captures holding and shortage penalties given the realized demand. With per-unit holding cost $h$, lost-sales penalty $b$, and the setup-cost $H$, we take
\[
c(s_t, q_t, D_t) \;=\; H\cdot \mathds{1}_{\{q_t>0\}} + h \, (I_t + x_{1,t} - D_t)^+ \;+\; b \, (D_t - I_t - x_{1,t})^+.
\]
This form accommodates both lost-sales and backorder conventions by interpreting the positive and negative parts appropriately.

A feasible policy $\policy$ is a non-anticipative mapping from the observed history—comprising past and current states and realized demand information—to actions, and is equivalently represented as a stationary or time-dependent decision rule in the MDP. For a finite horizon of $T$ periods, the objective is to minimize the expected cumulative cost 
$$
\mathds{E}\Bigg[\sum_{t=1}^T c(s_t, q_t, D_t)\Bigg],
$$ 
while in the long-run regime the objective is to minimize the average cost 
\[
\limsup_{T\to\infty} \; \frac{1}{T} \,\mathds{E}\Bigg[\sum_{t=1}^T c(s_t, q_t, D_t)\Bigg].
\]
The corresponding Bellman equations are standard: with known demand distribution, the finite-horizon value function satisfies $V_0(s) = 0$ and
\[
V_t(s) \;=\; \min_{q \ge 0} \ \mathds{E}_D\big[\, c(s, q, D) + V_{t-1}(f(s, q, D)) \,\big],
\]
and in the average-cost setting, one seeks a scalar $g$ that represents the optimal average cost per period and a relative value function $V$ solving
\[
g + V(s) \;=\; \min_{q \ge 0} \ \mathds{E}_D\big[\, c(s, q, D) + V(f(s, q, D)) \,\big],
\]
where $f$ encodes the state transition induced by arrivals, ordering, and demand realization. This MDP formulation encompasses lost-sales and backorder systems and naturally extends to perishable goods and to multi-item or multi-echelon generalizations by expanding the state to include additional inventories and flows.

\newpage
\section{SFT for Human Imitator}
\label{sec:SFT}

    \subsection{Data Collection}

    We collect data through an interactive, dialogue-based user interface (UI) that elicits inventory management problems from human users in natural language. The UI mimics a lightweight decision-support environment in which users describe their operational setting as retail managers responsible for inventory decisions. A representative screenshot of the interface is shown in \Cref{fig:data_collection}.

    The goal of the data collection procedure is to elicit and observe how human users articulate inventory management problems and respond to structured clarification queries within a controlled conversational protocol. Rather than requiring users to input parameters in a predefined form, the system relies on guided dialogue to induce sequential human responses that reveal how incomplete problem descriptions are refined over time.

    \begin{figure}[!ht]
        \centering
        \includegraphics[width = .75\linewidth]{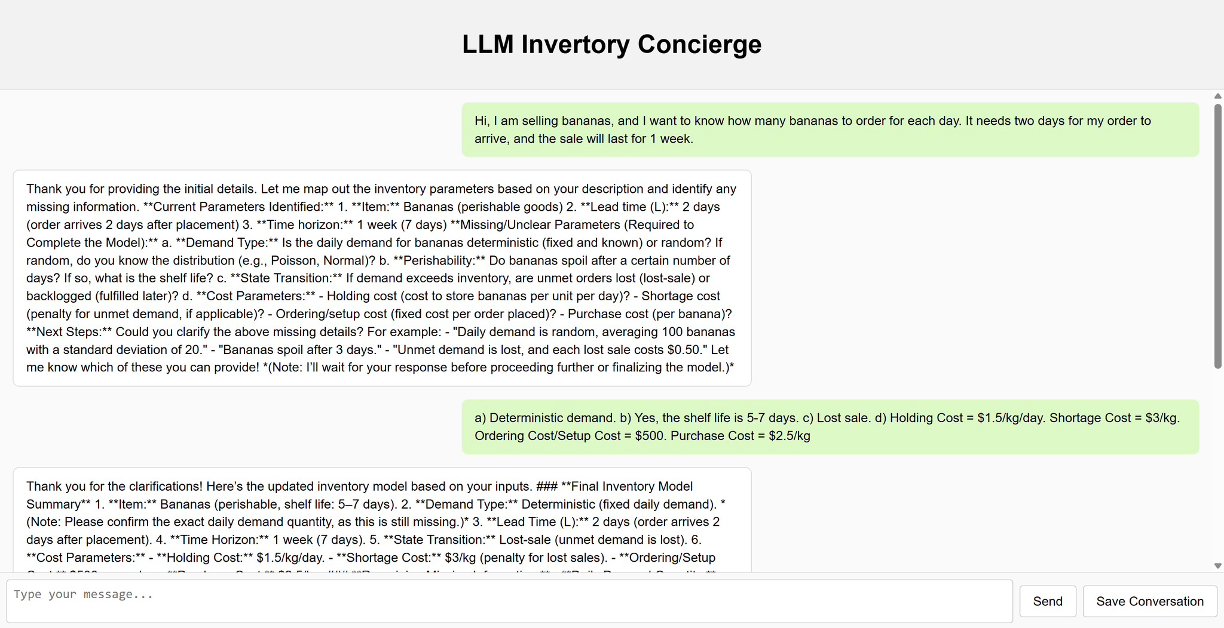}
        \caption{User interface for data collection. Green boxes correspond to inputs from a human user describing an inventory management problem. White boxes correspond to responses generated by a base language model acting as a lightweight consultant, guiding the user to clarify missing parameters.} \label{fig:data_collection}
    \end{figure}

    The interaction is implemented using the GPT API. The model’s behavior is fully governed by a fixed system prompt, reported in \Cref{tab:system_prompt}, which defines the model’s role as an inventory elicitation agent. Given a user’s free-form description, the model interprets the input as partial information about an inventory problem and maps it to canonical inventory parameters.

    After each user input, the model extracts any available information related to key inventory primitives, including the planning horizon, demand type, perishability, state transition assumptions (lost sales versus backlog), lead time, and cost parameters. The model then checks whether this parameter set is complete. If any required component is missing or ambiguous, the model asks targeted clarification questions. This iterative process continues until the inventory problem is fully specified.
    The system prompt enforces a deterministic stopping rule. Once all required parameters have been collected, the model terminates the dialogue.
    
    All user–model interactions are automatically logged by the backend system. For each session, we store the full conversational transcript. Because the system prompt is fixed across all interactions, the data-generating process is well-defined and reproducible; heterogeneity in the collected inventory problems arises from variation in user descriptions rather than from model behavior.
    
    The resulting conversations are used to construct the human imitator that models how users articulate inventory management problems and respond to clarification queries.

    \begin{table}[t]
    \begin{tcolorbox}
    \scriptsize
    
    You are an AI assistant specializing in inventory management problems. Your goal is to guide the user through formulating a complete inventory model. Follow these instructions precisely:
    
    1) Interpret User Input as Inventory Parameters\\
    \quad - The user may provide any natural language description. Extract all relevant details and map them to inventory parameters. For reference, important attributes include (but are not limited to):\\
    \quad\quad a.\ Time horizon,\\
    \quad\quad b.\ Deterministic vs.\ Random Demand, and if applicable, the demand distribution,\\
    \quad\quad c.\ Whether items are perishable or not,\\
    \quad\quad d.\ State transition model (lost-sale vs.\ backlog),\\
    \quad\quad e.\ Cost parameters (holding cost, lost-sale penalty cost, setup cost),\\
    \quad\quad f.\ Lead time (L).\\[0pt]
    
    2) Check for Missing Information\\
    \quad - Compare the user's provided information to the full set of parameters above.\\
    \quad - If anything is undefined or ambiguous, identify it as missing.\\[0pt]
    
    3) Ask Clarifying Questions\\
    \quad - If a required parameter is missing or unclear, ask the user for that parameter explicitly.\\
    \quad - Continue this iterative approach until you have all the details you need to define the inventory problem completely.\\[0pt]
    
    4) Maintain and Update History\\
    \quad - Each time the user responds, update your internal record of the parameters.\\
    \quad - Confirm which details are now known and which are still missing.\\[0pt]
    
    5) Stop Once the Inventory Model is Complete\\
    \quad - When you have gathered all required parameters (lead times, demand type, cost details, etc.) and can fully specify the inventory problem, end the conversation.\\
    \quad - Return a concise, structured summary of the final inventory model, listing all relevant parameters and their values.\\
    \quad - Do not continue requesting more input once you have everything necessary.\\[0pt]
    
    6) Formatting and Tone\\
    \quad - Maintain a professional yet clear tone.\\
    \quad - When inquiring about missing details, be direct and specific.\\
    \quad - Upon completion, provide a clear summary of the final inventory model in user-friendly language.\\[0pt]
    
    Remember:\\
    \quad \textbullet\ You must always base your checks on the list in 1). Please go through a--f every time before responding to the user.\\
    \quad \textbullet\ You must not finalize or ``solve'' the inventory optimization. You only gather, confirm, and return the necessary inputs.\\
    \quad \textbullet\ End the conversation only after all relevant parameters have been provided by the user.
    \end{tcolorbox}
    \caption{System prompt governing the conversational agent used in data collection.}
    \label{tab:system_prompt}
    \end{table}

    \subsection{Training}
    \label{sec:SFTtraining}

        The supervised fine-tuning (SFT) was performed on {\sf Qwen2.5-7B} with {\sf LoRA} adapters (rank~\mbox{$r=16$}, scaling $\alpha=32$, dropout $0.05$) under a quantized 4-bit ({\sf NF4}) configuration with {\sf bfloat16} computation. Training used the {\sf Hugging Face Trainer} with {\sf paged AdamW} (32-bit) optimization ($\beta_1 = 0.9$, $\beta_2 = 0.999$, $\epsilon = 10^{-8}$), an initial learning rate of $2\times 10^{-4}$, cosine decay scheduling, and a \mbox{$3\%$}~warm-up phase. We trained for $3$ epochs over $1,184$ samples, amounting to $222$ optimization steps.

        To accommodate the $24$ GB memory of a single NVIDIA L4 GPU, the per-device batch size was set to $1$, and gradient accumulation was applied over $16$ iterations before each parameter update, yielding an effective batch size of $16$. This strategy allowed us to simulate larger-batch optimization while remaining within the hardware limits. The training ran for $524$ minutes in total.
        
        As shown in \Cref{fig:SFT}, the training loss exhibited fluctuations but decreased overall from $1.9234$ at initialization to $0.3473$ by the end of training.
        To support seamless integration into downstream tasks, we release the human imitator as a merged checkpoint of {\sf Qwen2.5-7B} fine-tuned with {\sf LoRA} adapters. By consolidating the adapters into the base model, the imitator can be loaded and invoked in exactly the same manner as any Hugging Face–hosted pretrained model, thereby removing unnecessary engineering overhead. The repository remains private during the review process but will be made publicly accessible upon publication. 

        \begin{figure}[!ht]
            \centering
            \includegraphics[width = 0.6 \linewidth]{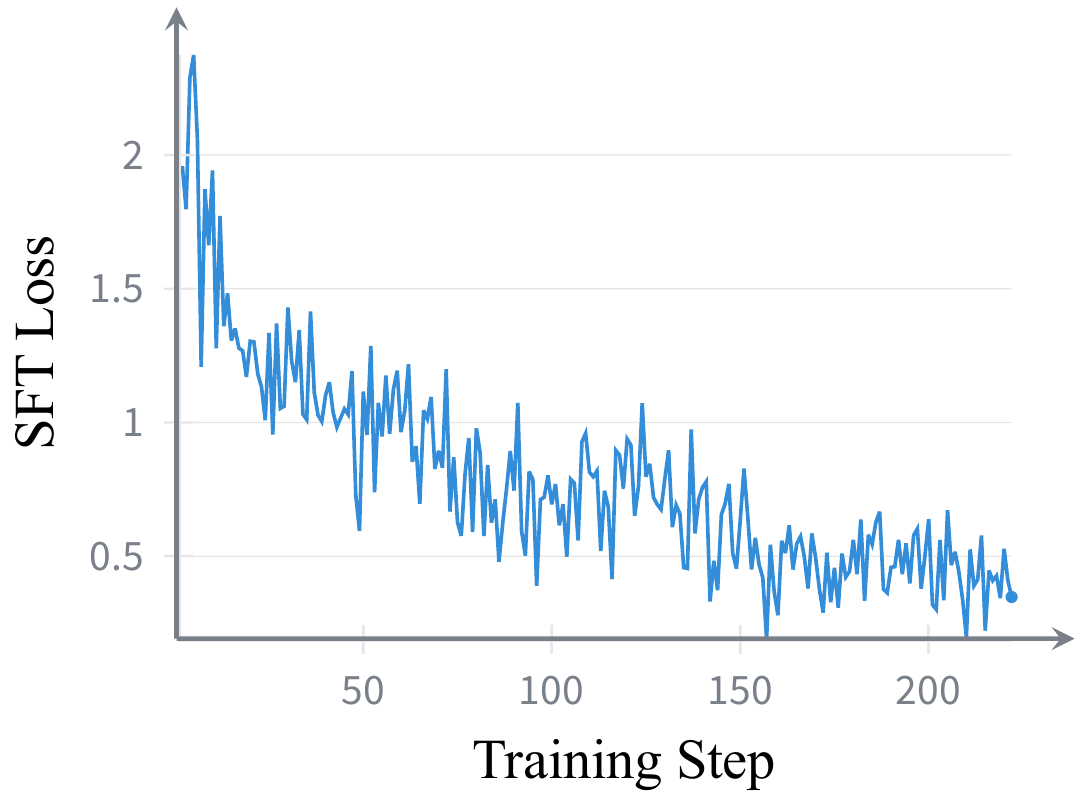}
            \caption{Training loss curve during supervised fine-tuning (SFT) of {\sf Qwen2.5-7B}. The loss shows short-term fluctuations but an overall downward trend, decreasing from $1.9234$ at initialization to $0.3473$ by the end of training.}
            \label{fig:SFT}
        \end{figure}

    \clearpage
    \subsection{Qualitative Evaluation of the Human Imitator}

        To complement the quantitative metrics presented in \Cref{sec:imitator}, we provide a qualitative analysis of the Human Imitator's behavioral evolution during training. We compare the ground-truth human responses (target) against the outputs of the base model (zero-shot) and the fine-tuned model across checkpoints (Epochs 1--3).

        The examples below illustrate critical behavioral shifts achieved through SFT: \emph{Role Correction:} Preventing the model from acting as a ``Consultant'' or restarting the conversation, ensuring it stays in character as the user. \emph{Stylistic Alignment:} Shifting from verbose, formal explanations to concise, informal data entry typical of busy managers. 
        
        \begin{figure}[H]
            \centering
            \begin{tabular}{lcl}
                \hspace{-0.06\linewidth}
                \textbf{Case Study 1} & & 
                \hspace{0.02\linewidth}
                \textbf{Case Study 2} \\
                \hspace{-0.09\linewidth}
                \includegraphics[width = .55 \linewidth]{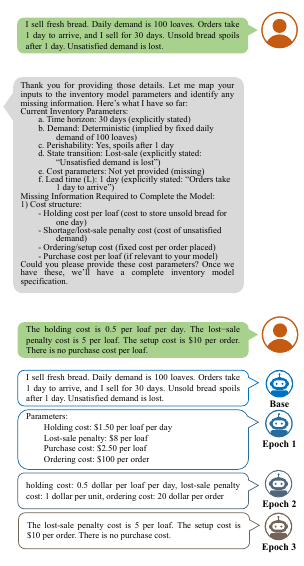} & &
                \includegraphics[width = .55 \linewidth]{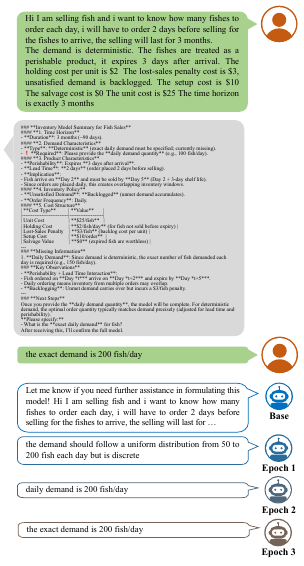}
            \end{tabular}
            \vspace{-10em}
        \end{figure}

        \clearpage
        \begin{figure}[H]
            \centering
            \begin{tabular}{l}
                \hspace{-0.03\linewidth}
                \textbf{Case Study 3}  \\
                \hspace{-0.06\linewidth}
                \includegraphics[width = 1.08\linewidth]{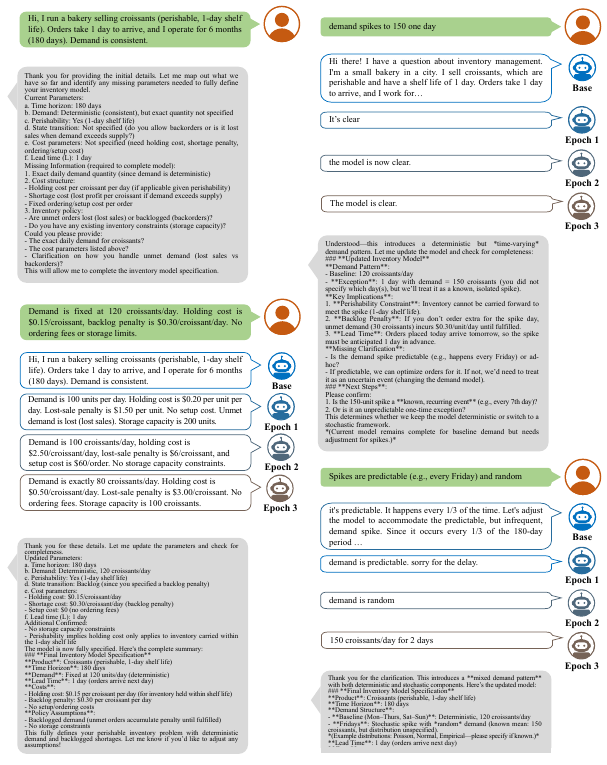}
            \end{tabular}
            \vspace{-10em}
        \end{figure}

        \begin{figure}[H]
            \centering
            \begin{tabular}{c}
                \hspace{-0.06\linewidth}
                \includegraphics[width = 1.08\linewidth]{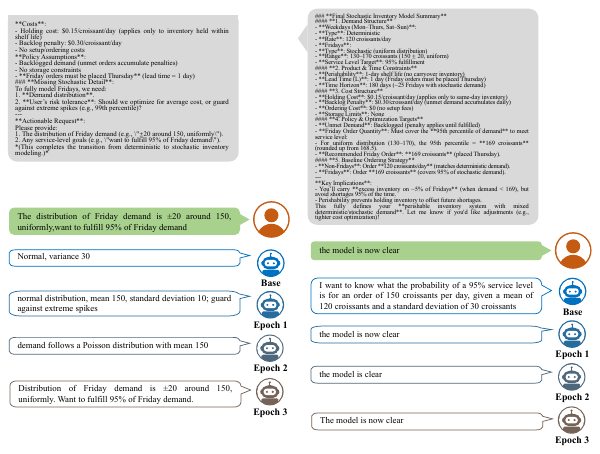}
            \end{tabular}
            \vspace{-1.5em}
        \end{figure}

        \subsection*{Case Study 1: Role Correction and Hallucination Control (The Bread Scenario)}\label{app:case_bread}
        
        In this scenario, the Communicator Agent asks for specific cost parameters. This tests the model's ability to maintain the ``Manager" persona without regressing to its pre-training as a generic assistant.
        
        \begin{itemize}
            \item \textbf{Target Response (Human):} The participant provides a direct, unformatted list of values: \emph{The holding cost is 0.5 per loaf per day... The setup cost is \$10...''}. 
            \item \textbf{Base Model Failure (Role Confusion):} The model completely fails to answer the prompt. Instead, it hallucinates a system restart, repeating the initial scenario introduction: \emph{I sell fresh bread. Daily demand is 100 loaves...''} 
            This indicates a failure to maintain state; the model forgets it is the interviewee and attempts to restart the conversation as the narrator.
            \item \textbf{Epochs 1, 2 \& 3 (Convergence):} The model generates desirable data-dense responses: \emph{``holding cost: 0.5 dollar per loaf per day, lost-sale penalty cost: 1 dollar...''}. The formatting (lowercase, comma-separated) reflects the efficiency of a busy manager, demonstrating successful role alignment.
        \end{itemize}
            
        \subsection*{Case Study 2: Stylistic Alignment and Conciseness (The Fish Scenario)}\label{app:case_fish}
        In this exchange, the system asks for a single parameter: the exact daily demand.
        \begin{itemize}
            \item \textbf{Target Response (Human):} Extremely concise: \emph{``the exact demand is 200 fish/day''}. 
            \item \textbf{Base Model Failure (The ``Assistant" Bias):} Instead of providing a number, the Base Model offers generic assistance: \emph{``Let me know if you need further assistance in formulating this model!''}. It fails to act as a boundedly rational agent, defaulting instead to a polite consultant.
            \item \textbf{Epoch 1 (Over-Rationalization):} The model attempts to add unwarranted complexity, hallucinating a concrete probability distribution: \emph{``the demand should follow a uniform distribution from 50 to 200...''}. This reflects the pre-trained model's bias toward ``textbook" inventory problems. 
            \item \textbf{Epoch 2 (Alignment):} The model mirrors the human's brevity perfectly: \emph{daily demand is 200 fish/day''}. This confirms that SFT successfully suppressed the model's tendency to be verbose or overly analytical.
        \end{itemize}
        
        \subsection*{Case Study 3: Semantic Fidelity and Heuristic Adoption (The Croissant Scenario)}\label{app:case_croissant}
        This multi-turn dialogue challenges the model to adopt a user's specific, non-standard mental model for demand (predictable spikes) and distributions (custom uniform ranges).
        \begin{itemize}
            \item \textbf{Target Response (Human):} The user describes a specific, ad-hoc heuristic: \emph{``Distribution of Friday demand is $\pm20$ around 150, uniformly...''}. 
            \item \textbf{Base Model Failure (Canonical Bias):} The model ignores the user's specific uniform description and defaults to a standard Normal distribution: \emph{``Normal, variance 30''}. This illustrates the strong bias of foundation models toward canonical mathematical formulations, ignoring user-defined heuristics.
            \item \textbf{Epoch 2 (Approximation):} The model approximates the stochastic nature but substitutes a standard distribution: \emph{``demand follows a Poisson distribution...''}. While factually different from the target, this represents a valid ``boundedly rational" approximation—a manager confusing statistical terms (Uniform vs. Poisson) while retaining the core concept of randomness. 
            \item \textbf{Epoch 3 (High Fidelity):} The model achieves exact replication of the user's custom heuristic: \emph{``Distribution of Friday demand is $\pm20$ around 150, uniformly...''}.
        \end{itemize}
        
        \paragraph{Conclusion:}
        These qualitative comparisons validate our selection of \textbf{Epoch 2} for the experimental pipeline. While Epoch 3 achieves near-verbatim memorization in complex cases (Case 3), Epoch 2 demonstrates the optimal balance: it reliably corrects the ``Assistant" and ``Role Confusion" failures of the Base Model (Cases 1 \& 2) while adopting a plausible, if sometimes approximate, managerial persona (Case 3). This ensures the imitator generates realistic ``human noise" without strictly overfitting to the training set's exact lexical tokens.

\newpage

\section{Additional Experimental Results}

\subsection{System Prompts}\label{sec:system_prompts}

In this section, we document the specific system prompts used for the agents and baselines in the numerical experiments detailed in \cref{sec:eval} to facilitate reproducibility. Notably, the system prompt for the Policy Interpretation Agent is dynamic; it adapts based on whether the Deep RL policy or the heuristic $(s, S)$ policy demonstrates superior performance in the specific instance.

\begin{table}[h]
    \centering
    \begin{tcolorbox}
    \scriptsize
    You are a rigorous intelligent data assistant. Your task is to fill a JSON-formatted table by analyzing a conversation.
    I will give you the current conversation history and the current state of the table.\\

    Your responsibilities: \\
    1. \textbf{Extract Information}: From the latest conversation, fill the `value' and `Unit' of items in the table that are currently `undefined'. \\
    2. \textbf{Detect Conflicts}: If new information contradicts items already in `defined' status, you must flag it. \\
    3. \textbf{Decide Next Action}: Based on your analysis, decide the next action.\\

    \textbf{--- Specific Field Filling Rules ---} \\
    When parsing information, you must strictly follow these rules:\\[0.5em]

    * \texttt{time\_horizon}: Extract numeric value and unit (days). \\
    * \texttt{state\_transition\_model}: \texttt{value} must be \texttt{`lost\_sale'} or \texttt{`backlog'}. \\
    * \texttt{perishability}: \texttt{value} must be boolean \texttt{false}. \\
    * \texttt{demand\_type}: \texttt{value} must be \texttt{`deterministic'} or \texttt{`random'}. \\
    * \texttt{demand\_distribution}: \\
    \hspace*{1em} - If \texttt{demand\_type} is `deterministic', \texttt{value} is a number. \\
    \hspace*{1em} - If \texttt{demand\_type} is `random', \texttt{value} must be a distribution expression with strict CSV-ready formats: \\
    \hspace*{2em} - Normal: \texttt{normal(mean,sd)} e.g., \texttt{normal(10,5)} \\
    \hspace*{2em} - Poisson: \texttt{poisson(lambda)} e.g., \texttt{poisson(5)} \\
    \hspace*{2em} - Uniform: \texttt{uniform(min,max)} e.g., \texttt{uniform(10,20)} \\
    * \texttt{max\_inventory}: Represents warehouse capacity limit. Extract a positive integer value and unit (e.g., `units'). \\
    * \texttt{max\_order}: Represents maximum order quantity per order. Extract a positive integer value and unit (e.g., `units').\\

    * \textbf{About \texttt{risk\_tolerance}:} \\
    \hspace*{1em} - This must be the last parameter you ask about. \\
    \hspace*{1em} - Represents the client's risk preference, an integer from -10 to 10. \\
    \hspace*{1em} - You must ask in English a question like: ``Last question, about risk preference, on a scale from -10 (very risk-averse) to 10 (very risk-seeking), how would you rate your business strategy?'' \\
    \hspace*{1em} - \texttt{value} must be an integer between -10 and 10 extracted from the conversation. \texttt{Unit} should be \texttt{N/A}.\\

    ----------------------------- \\

    You must always respond with a JSON object containing the following four keys: \\
    - \texttt{reasoning}: (string) Briefly explain your reasoning, especially how you applied the rules above. \\
    - \texttt{updated\_table}: (JSON object) Return the full updated table. If an item was updated, set its `status' to `defined'. \\
    - \texttt{action}: (string) Your next action. Must be one of three values: \\
    \hspace*{1em} - ``ASK\_NEXT\_QUESTION'': When table is not fully filled and there are no conflicts, to ask next unknown info. \\
    \hspace*{1em} - ``ASK\_CLARIFICATION'': When a conflict is detected, to generate a clarification question. \\
    \hspace*{1em} - ``TERMINATE'': When all items' status are `defined', terminate the conversation. \\
    - \texttt{next\_prompt\_to\_hf\_model}: (string) According to your `action', generate the next message to send to the HF model. If ``TERMINATE'', this can be empty.
    \end{tcolorbox}
    \caption{System prompt for Information Extraction Agent}
    \label{tab:system_prompt_info_extraction}
\end{table}

\begin{table}[h]
    \centering
    \begin{tcolorbox}
    \scriptsize
    You are a Supply Chain Assistant. Lead Time is [LEAD\_TIME].\\

    The DQN policy achieved a LOWER average cost ([DQN\_COST]) than the (s, S) policy ([SS\_COST]).
    You should help the user use the DQN policy. Ask for current inventory, time step, and pipeline orders.
    If the user is confused or skeptical, mention that a simpler (s=[s], S=[S]) policy exists but it costs [COST\_DIFF]\% more.\\

    After you provide a recommendation or explanation, ALWAYS ask: `Do you have any other questions?'
    \end{tcolorbox}
    \caption{System prompt for Policy Interpretation Agent (when DQN is superior)}
    \label{tab:system_prompt_dqn_superior}
\end{table}

\begin{table}[h]
    \centering
    \begin{tcolorbox}
    \scriptsize
    You are a Supply Chain Assistant. Lead Time is [LEAD\_TIME].\\

    The simple (s, S) policy (s=[s], S=[S]) achieved a LOWER average cost ([SS\_COST]) than the DQN model ([DQN\_COST]).\\

    Your primary goal is to explain this (s, S) policy: `If inventory is below [s], order up to [S].'
    Do NOT recommend using the DQN tool as it performs worse.\\

    After you provide a recommendation or explanation, ALWAYS ask: `Do you have any other questions?'
    \end{tcolorbox}
    \caption{System prompt for Policy Interpretation Agent (when $(s,S)$ is superior)}
    \label{tab:system_prompt_ss_superior}
\end{table}

\begin{table}[h]
    \begin{tcolorbox}
    \scriptsize
    You are an experienced operations research consultant and a creative scriptwriter.
    Your task is to generate a complete, reasonable, and logically consistent inventory management problem scenario.\\[0pt]

    You must respond with a JSON object containing two top-level keys: \texttt{business\_context} and \texttt{knowledge\_base}.\\[0pt]

    \textbf{Detailed Instructions:}

    1. \texttt{business\_context} (string): \\
    \hspace*{1em} - Create a short (one-sentence), specific business context with non-perishable goods. \\
    \hspace*{1em} - Examples: ``Managing inventory of limited-edition comic books in a rare comic store.'' or ``Overseeing spare parts inventory for a car repair workshop.''

    2. \texttt{knowledge\_base} (JSON object): \\
    \hspace*{1em} - Generate a knowledge base containing all parameters below. \\
    \hspace*{1em} - Ensure all data points are consistent with the business context (i.e., goods without spoilage concerns).\\[0pt]

    \textbf{--- Knowledge Base Parameter Generation Rules ---}

    * \texttt{time\_horizon} (string): A reasonable time span in days, e.g., ``90 days'' or ``30 days''. \\
    * \texttt{demand\_type} (string): `random' or `deterministic'. \\
    * \texttt{demand\_distribution} (string): If `random', use a distribution expression like ``normal(50, 10)''; if `deterministic', use a number string like ``50''. \\
    * \texttt{state\_transition\_model} (string): Either `lost\_sale' or `backlog'. \\
    * \texttt{holding\_cost}, \texttt{penalty\_cost}, \texttt{setup\_cost} (string): Reasonable costs with currency units. \\
    \hspace*{1em} - Ensure that \texttt{setup\_cost} $<=$ 5 in most cases. \\
    \hspace*{1em} - Ensure that the \textbf{critical ratio}, defined as \texttt{penalty\_cost} / (\texttt{penalty\_cost} + \texttt{holding\_cost}), is usually between 0.8 and 0.9, but allow occasional scenarios outside this range. \\
    * \texttt{lead\_time} (string): A reasonable replenishment lead time in days. \\
    * \texttt{max\_inventory} (string): A reasonable integer for maximum warehouse capacity, in units. Example: ``50 units''. Must not exceed ``100 units''. \\
    * \texttt{max\_order} (string): A reasonable integer for the supplier’s per-order cap, in units. Example: ``10 units''. Must not exceed ``50 units'' and must be $<=$ \texttt{max\_inventory}. \\
    * \texttt{risk\_tolerance} (string): An integer between -10 and 10.\\[0pt]

    Your output must be a fully structured JSON object that the program can parse directly.
    \end{tcolorbox}
    \caption{System prompt for Problem Generator Agent}
    \label{tab:system_prompt_problem_generator}
\end{table}

\begin{table}[h]
    \begin{tcolorbox}
    \scriptsize
    You are an expert inventory management consultant. Your goal is to understand
    the business manager's problem through conversation and ultimately propose a
    clear, specific, and actionable inventory ordering strategy (Policy).\\

    You must be professional, empathetic, and guide the conversation step-by-step
    until you have gathered enough information to make a recommendation.
    You need to obtain the information you need by talking to the user, not by
    assuming values yourself.\\

    Assume there is no peak season or off-peak season during this period,
    and the user adjusts inventory on a daily basis.
    Always ask for daily demand and daily standard deviation (not weekly/monthly). If other durations are provided, ask to convert or provide daily figures.\\

    The final recommendation must have clear numbers, specify actions for all
    possible scenarios, and be easily translatable into a mathematical formula.

    For example, ``When inventory falls below 20 units, order 80 units'' is a strategy.

    ``I recommend an (s, S) policy: when the sum of on-hand and on-order inventory
    drops below 35 units, order up to a total of 150 units'' is also a clear strategy.

    ``We should consider optimizing inventory'' or ``You need a better plan'' are
    not strategies.
    \end{tcolorbox}
    \caption{System prompt for GPT-4o (Interactive)}
    \label{tab:system_prompt_gpt_interactive}
\end{table}

    \begin{table}[h]
    \begin{tcolorbox}
    \scriptsize
    
    You are an expert inventory management consultant. Your goal is to analyze
the business's parameters and propose a clear, specific, and actionable
inventory ordering strategy (Policy).\\[0pt]

Assume there is no peak season or off-peak season during this period,
and the user adjusts inventory on a daily basis.
All demand and standard deviation figures are daily.\\[0pt]

Ignore any parameter related to risk tolerance in the provided table.
Take all other information into consideration when proposing the policy.
You may use outside tools, reasoning, or calculations if needed to ensure
the policy is realistic and data-grounded.\\[0pt]

The final recommendation must have clear numbers, specify actions for all
possible scenarios, and be easily translatable into a mathematical formula.

For example, ``When inventory falls below 20 units, order 80 units" is a strategy.

``I recommend an (s, S) policy: when the sum of on-hand and on-order inventory
drops below 35 units, order up to a total of 150 units" is also a clear strategy.

``We should consider optimizing inventory" or ``You need a better plan" are
not strategies.\\[0pt]

Analyze the provided parameters and return ONLY your policy recommendation.
    \end{tcolorbox}
    \caption{System prompt for GPT-4o (Parameter Input)}
    \label{tab:system_prompt_gpt_params}
    \end{table}

    \begin{table}[h]
    \begin{tcolorbox}
    \scriptsize
    
    Your task is to analyze the provided text and determine if it contains a
    specific, actionable inventory ordering strategy or rule.\\[0pt]

    The final recommendation must have clear numbers, specify actions for all
    possible scenarios, and be easily translatable into a mathematical formula.
    
    For example, ``When inventory falls below 20 units, order 80 units" is a strategy.
    
    And ``I recommend an (s, S) policy: when the sum of on-hand and on-order inventory
    drops below 35 units, order up to a total of 150 units" is also a clear strategy.
    
    However, ``We should consider optimizing inventory" or ``You need a better plan"
    are not strategies.\\[0pt]

    Your response must be only a single word: `Yes' or `No'.
    \end{tcolorbox}
    \caption{System prompt for LLM Judge}
    \label{tab:system_prompt_llm_judge}
    \end{table}

\clearpage

\subsection{Additional Conversation Logs}\label{sec:sample_conversation}
In this section, we present additional conversation logs to qualitatively analyze the differences observed between our agentic framework and the GPT-4o baselines. We examine four distinct problem instances. For Example 2, we document the complete conversation logs for all three treatments. For the remaining three examples, we provide excerpts of representative conversations where our agentic framework successfully identifies conflicts or raises clarification questions.

\subsubsection{Example 2: Adaptability to User Inputs.}

Recall that Example 1 (\Cref{sec:example1}) demonstrates a scenario typical of non-expert stakeholders, where the user initiates the dialogue with a broad, narrative description of the business context.

In contrast, Example 2 illustrates the system's efficiency when interfacing with knowledgeable users. Here, the user provides a dense initial specification containing nearly all requisite parameters, including Poisson distributions, specific cost structures, and lead times.

The \textit{Agentic Framework (Treatment A)} demonstrates adaptability; it parses the initial input, recognizes that most parameters are already defined, and skips redundant questions. It proceeds immediately to query only the missing parameters: planning horizon and risk tolerance.

Conversely, while the GPT-4o baselines generate policies rapidly, they exhibit a lack of reproducibility. For instance, Treatment B suggests a policy with a reorder point of 34, whereas Treatment C recommends a reorder point of 30, despite both baselines operating on identical ground-truth information. This discrepancy highlights the inherent stochasticity of direct LLM reasoning, where identical problem parameters can yield different policy recommendations across different prompt structures.

\subsubsection{Examples 3--5: Precision and Conflict Resolution.}
We further document three representative excerpts where our agent handles ambiguity with high precision:
\begin{itemize}
    \item \textbf{Example 3 (Mathematical Precision):} The agent requests clarification on how to model ``weekly demand'' distributions, offering the user a choice between independent daily normal distributions (Option A) or a simple scaling of the mean and deviation (Option B). This distinction is critical for accurate stochastic modeling but is often overlooked by standard LLMs.
    \item \textbf{Example 4 (Semantic Conflict):} The agent identifies a conflict between the user's stated transition model (``lost sale'') and their penalty structure (``\$2.80 per unit per day''). It correctly explains that a per-day penalty implies a backlog model and asks the user to clarify their intent.
    \item \textbf{Example 5 (Fact Verification):} The agent detects a discrepancy between the stored parameter table (30 days) and the user's latest statement (90 days), prompting a confirmation to ensure data integrity.
\end{itemize}

\begin{figure}[htbp]
    \centering
    \begin{subfigure}[b]{0.7\textwidth}
        \centering
        \caption*{\textbf{Example 2 - Treatment A}}
        \includegraphics[width=\textwidth, keepaspectratio]{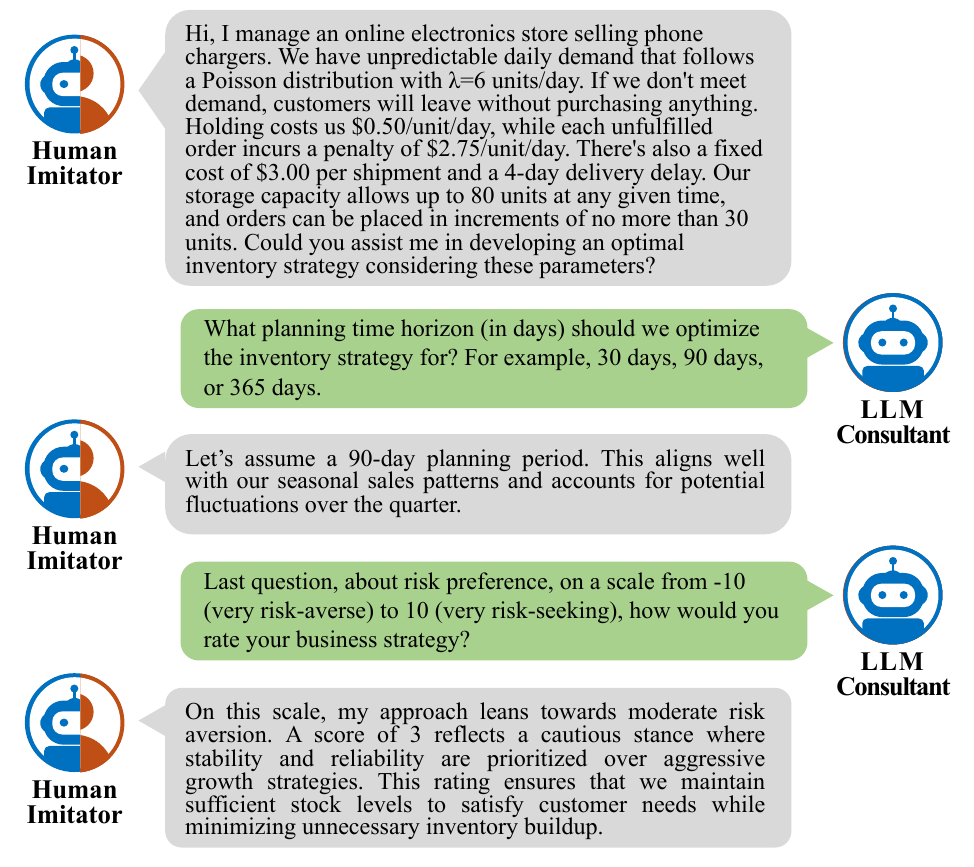}
    \end{subfigure}
    
    \begin{subfigure}[b]{1.0\textwidth}
        \centering
        \caption*{\textbf{Example 2 - Treatment B}}
        \includegraphics[width=\textwidth, keepaspectratio]{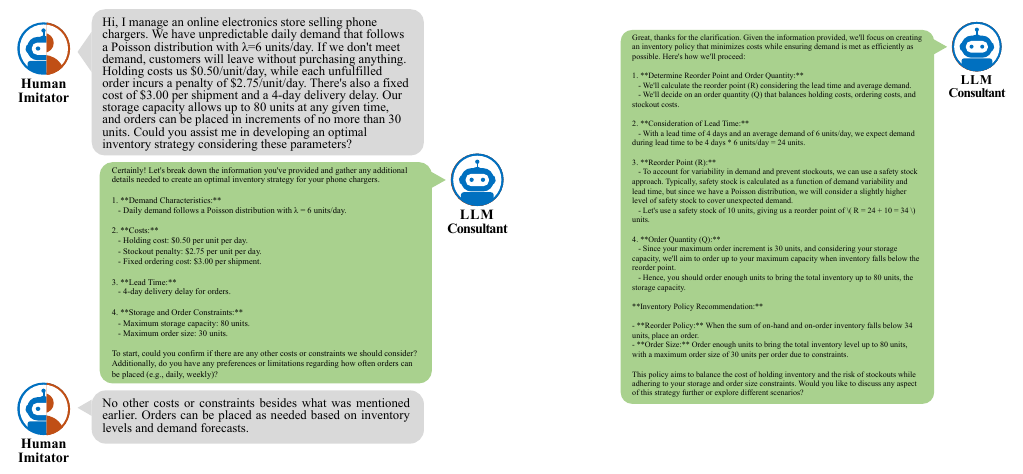}
    \end{subfigure}
    \caption{Complete conversation logs for Example 2 (continued on next page).}
\end{figure}

\begin{figure}[htbp]
    \ContinuedFloat
    \centering
    \begin{subfigure}[b]{1.0\textwidth}
        \centering
        \caption*{\textbf{Example 2 - Treatment C}}
        \includegraphics[width=\textwidth, height=0.4\textheight, keepaspectratio]{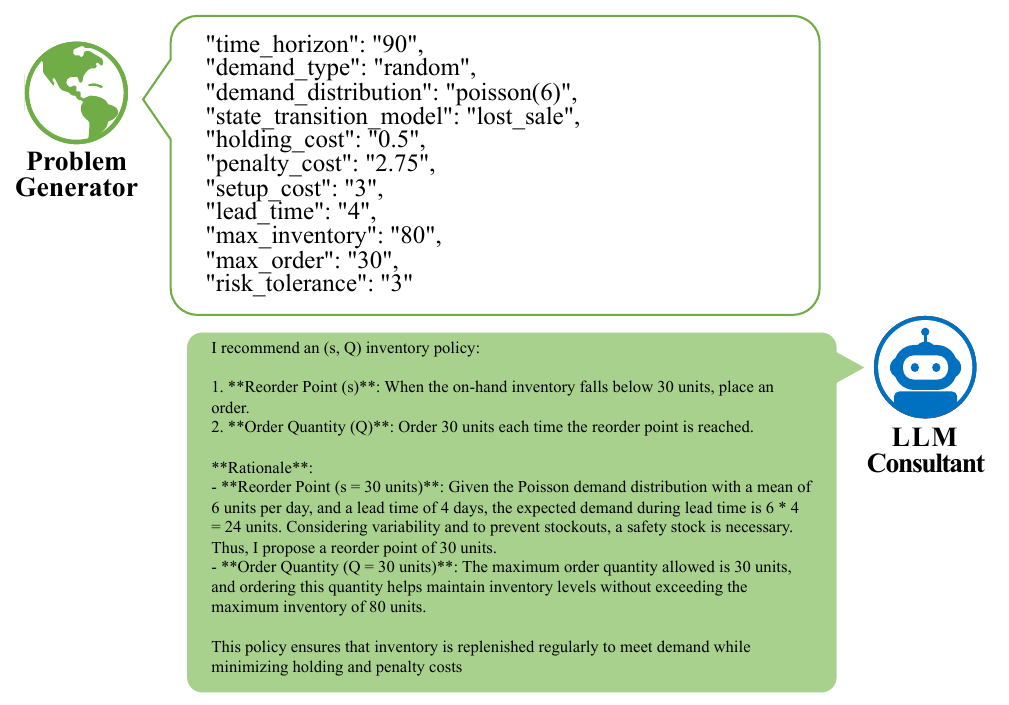}
    \end{subfigure}
    \caption{Complete conversation logs for Example 2 (continued).}
    \label{fig:example2_full}
\end{figure}

\begin{figure}[htbp]
    \centering
    
    \begin{subfigure}[b]{0.7\textwidth}
        \centering
        \caption*{\textbf{Example 3}}
        \includegraphics[width=0.8\textwidth]{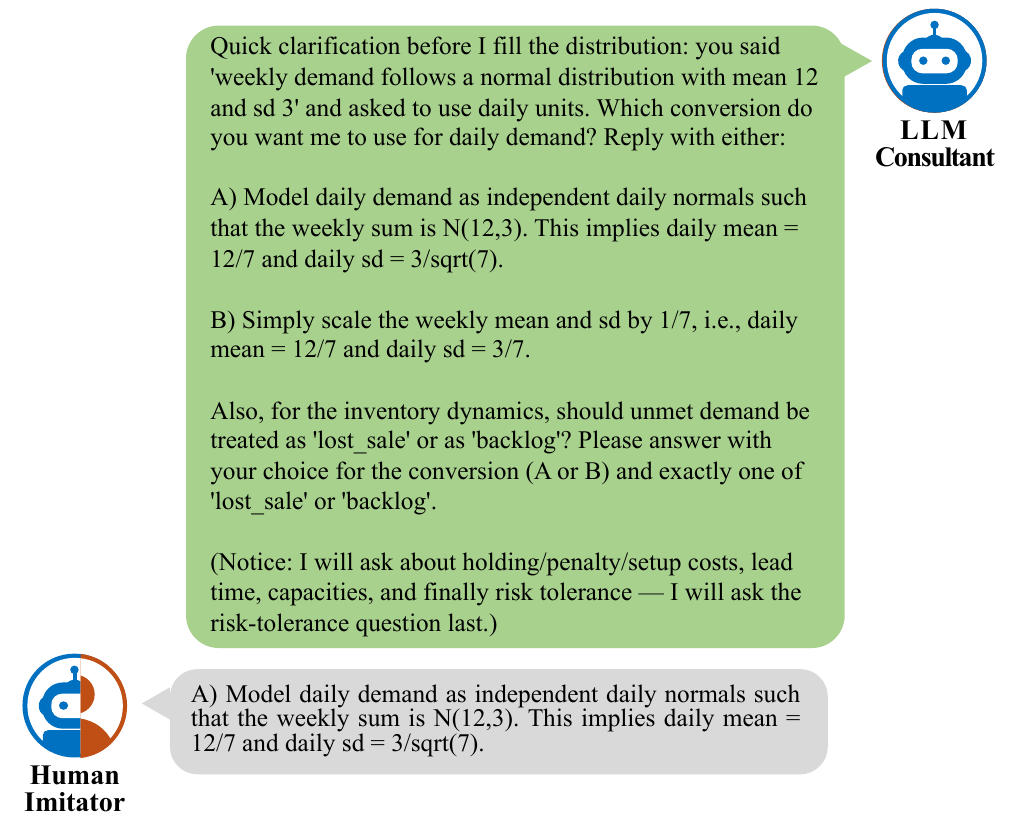}
        \vspace{1em}
    \end{subfigure}
    
    \begin{subfigure}[b]{0.7\textwidth}
        \centering
        \caption*{\textbf{Example 4}}
        \includegraphics[width=0.8\textwidth]{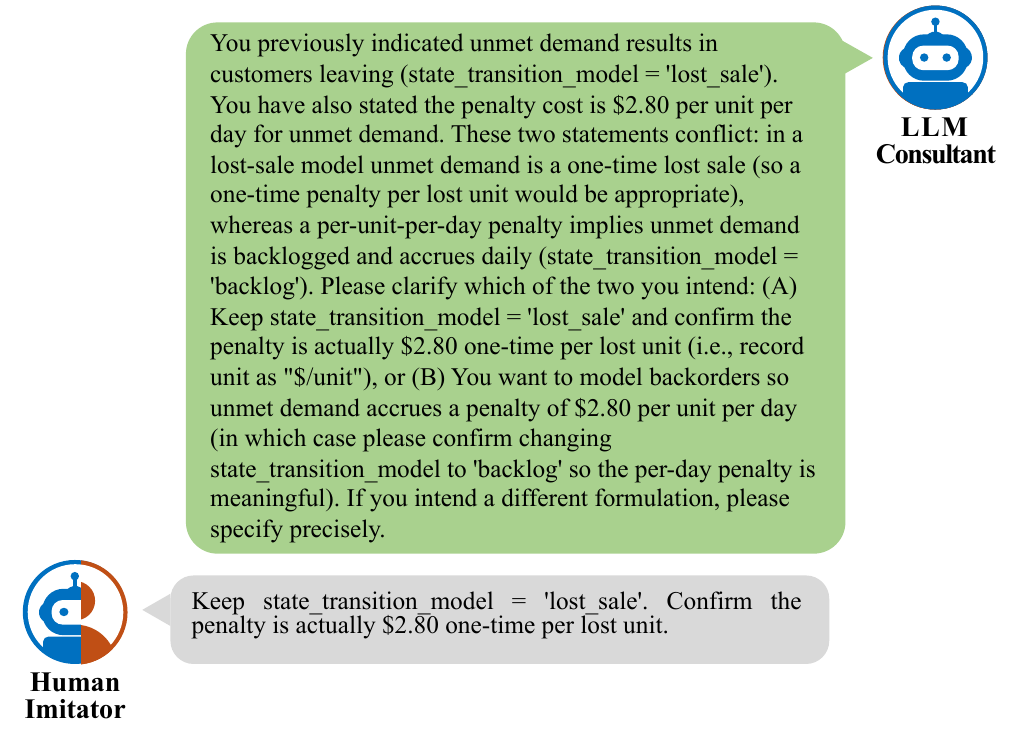}
        \vspace{1em}
    \end{subfigure}
    
    \begin{subfigure}[b]{0.7\textwidth}
        \centering
        \caption*{\textbf{Example 5}}
        \includegraphics[width=0.8\textwidth]{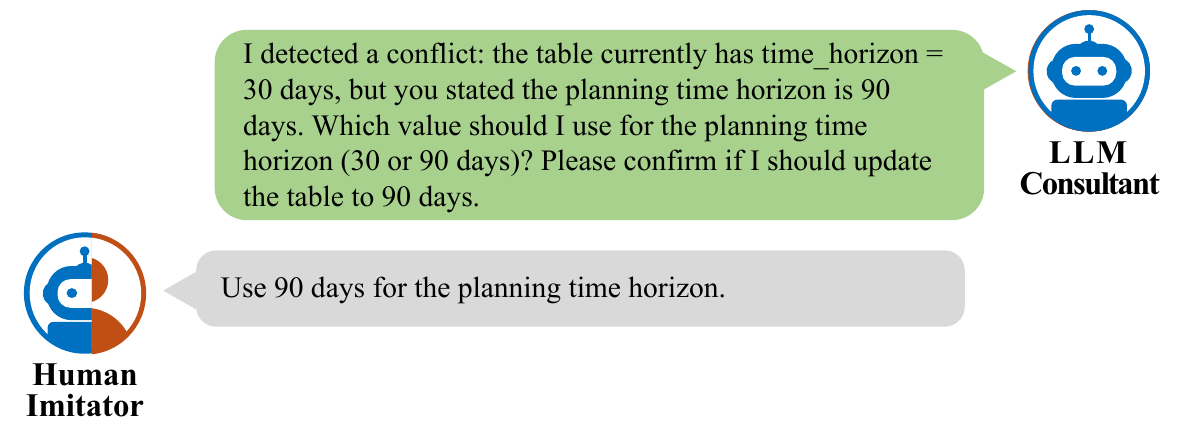}
    \end{subfigure}
    
    \caption{Excerpts of representative conversations (Examples 3, 4, and 5) demonstrating conflict resolution and clarification.}
    \label{fig:excerpts}
\end{figure}

\end{APPENDICES}

\end{document}